\documentclass[runningheads]{llncs}
\usepackage{graphicx}
\usepackage{amsmath,amssymb} %
\usepackage{color}
\usepackage{array}
\usepackage{multirow}
\usepackage{dcolumn}
\usepackage[width=122mm,left=12mm,paperwidth=146mm,height=193mm,top=12mm,paperheight=217mm]{geometry}
\begin{document}
\pagestyle{headings}

\mainmatter

\title{VConv-DAE: Deep Volumetric Shape Learning Without Object Labels}

\newcolumntype{d}[1]{D{.}{.}{#1}}
\author{Abhishek Sharma$^{1}$, Oliver Grau$^{2}$, Mario Fritz$^{3}$}
\institute{$^{1}$Intel Visual Computing Institute $^{2}$Intel $^{3}$Max Planck Institute for Informatics}
\authorrunning{ }

\maketitle

\begin{abstract}
With the advent of affordable depth sensors,  3D capture becomes more and more ubiquitous and already has made its way into commercial products. Yet, capturing the geometry or complete shapes of everyday objects using scanning devices (e.g. Kinect) still comes with several  challenges that result in noise or even incomplete shapes.

Recent success in deep learning has shown how to learn complex shape distributions in a data-driven way from large scale 3D CAD Model collections and to utilize them for 3D processing on volumetric representations and thereby circumventing problems of topology and tessellation. Prior work has shown encouraging results on problems ranging from shape completion to recognition. We provide an analysis of such approaches and discover that training as well as the resulting representation are strongly and unnecessarily tied to the notion of object labels.
Thus, we propose a full convolutional volumetric auto encoder that learns volumetric representation from noisy data by estimating the voxel occupancy grids. The proposed method outperforms prior work on challenging tasks like denoising and shape completion. We also show that the obtained deep embedding gives competitive performance when used for classification and promising results for shape interpolation.
\keywords{ Denoising auto-encoder, 3D deep learning, shape completion, shape blending}
\end{abstract}

\section{Introduction}
Despite the recent advances in 3D scanning technology, acquiring 3D geometry or shape of an object is a challenging task. Scanning devices such as Kinect are very useful but  suffer from problems such as sensor noise, occlusion, complete failure modes (e.g. dark surfaces and gracing angles). Incomplete geometry poses severe challenges for a range of application such as  interaction with the environment in Virtual Reality or Augmented Reality scenarios, planning for robotic interaction or 3D print and manufacturing.

 To overcome some of these difficulties, there is  a large body of work on fusing multiple scans into a single 3D model \cite{kin_fusion}. While the surface reconstruction is impressive in many scenarios, acquiring geometry from multiple viewpoint can be infeasible in some situations. For example, failure modes of the sensor will not be resolved and some viewing angles might simply be not easily accessible e.g. for a bed or cupboard placed against a wall or chairs occluded by tables.

There also has been significant research on analyzing 3D CAD model collections of everyday objects. Most of this work ~\cite{ckgk-prabm-11,KCKK12} use an assembly-based approach to build part based models of shapes. Thus, these methods rely on part annotations and can not model variations of large scale shape collections across classes. Contrary to this approach,  Wu et al. (Shapenet~\cite{Shape_net}) propose a first attempt to apply  deep learning to this task and learn the complex shape distributions in a data driven way from raw 3D data. It achieves  generative capability by formulating a probabilistic model over the voxel grid and labels. Despite the impressive and promising results of such Deep belief nets~\cite{hinton2006fast,lee2011unsupervised}, these models can be challenging to train. While they show encouraging results on challenging task of  shape completion, there is no quantitative evaluation. Furthermore, it requires costly sampling techniques for test time inference, which severely limits the range of future 3D Deep Learning applications.

While deep learning has made remarkable progress in computer vision problems with powerful hierarchical feature learning, unsupervised feature learning remains a future challenge that is only slowly getting more traction. Labels are even more expensive to obtain for 3D data such as point cloud. Recently, Lai et al.~\cite{lai_icra14} propose a sparse coding method for learning  hierarchical feature representations over  point cloud data. However, their approach is based on dictionary learning which is generally slower and less scalable than convolution based models .
Our work also falls in this line of work and aims at bringing the success of deep and unsupervised feature learning to 3D representations.

\vspace{0.1cm}
\noindent
To this end, we make the following contributions:
\begin{itemize}
 \item We propose a fully convolutional volumetric auto-encoder which, to our knowledge, is the first attempt to learn a deep embedding of object shapes in an unsupervised fashion.
 \item  Our method outperforms previous supervised approach of shapenet ~\cite{Shape_net} on denoising and shape completion task while it obtains competitive results on shape classification. Furthermore, shape interpolation on the learned embedding space shows promising results.
 \item  We provide an extensive quantitative evaluation protocol for task of shape completion that is essential to compare and evaluate the generative capabilities of deep learning when obtaining ground truth of real world data is challenging.
\item Our method is trained from scratch and end to end thus circumventing the training issues of previous work, shapenet~\cite{Shape_net}, such as layer wise pre-training. At test time, our method is at least two orders of magnitude faster than shapenet.
\end{itemize}

\section{Related Work}
\paragraph{Part and Symmetry based Shape Synthesis.}
Prior  work~\cite{ckgk-prabm-11,KCKK12} uses an assembly-based approach to build deformable part-based models based on CAD models. There is also work that detect the symmetry in point cloud data and use it to complete the partial or noisy reconstruction. A comprehensive survey of such techniques is covered in  Mitra \emph{et al.}\cite{mpwc_symmSurvey_12}. Huang  \emph{et al.}\cite{HM16} learns to predict  procedural model parameters for shape synthesis given  a 2D sketch using CNN.  However, part and symmetry based methods are typically class specific and require part annotations which are expensive. In contrast, our work does not require additional supervision in the form of parts, symmetry, multi-view images or their correspondence to 3D data.

\paragraph{Deep learning for 3D data.}
ShapeNet~\cite{Shape_net} is the first work that applied deep learning to learn the 3D representation on large scale CAD model database. Apart from recognition, it also desires capability of shape completion. It builds a generative model with convolutional RBM~\cite{hinton2006fast,lee2011unsupervised} by learning a probability distribution over class labels and voxel grid. The learned model is then fine tuned for the task of shape completion. Following the success of Shapenet, there have been recent work that improves the recognition results on 3D data~\cite{Maturana_2015_7900}, uses 3D-2D(multi-view images) correspondence to improve shape completion (repairing) results~\cite{field_repairing_CVPR16,3DDmind} propose intrinsic CNN~\cite{BC16,BB16} for 3D data or  learn correspondence between two surfaces(depth map)~\cite{WL16}.
Our work is mainly inspired by Shapenet in the functionality but differs in methodology. In particular, our network is trained completely unsupervised and discovers useful visual representations without the use of explicitly curated labels. By learning to predict missing voxels from input voxels, our model ends up learning an embedding that is useful for both classification as well as interpolation.

\paragraph{Denoising Auto-Encoders.} Our network architecture is inspired by
DAE~\cite{Vincent08,vincent2010stacked} main principle that predicting any subset of variables from the rest is a sufficient condition for completely capturing the joint distribution between a set of variables.
  In order to share weights for stationary distributions such as they occur in images, convolutional auto-encoders have been proposed \cite{masci2011stacked}. Our model differs with such architecture as it is not stacked and learned end to end without any layer fine-tuning or pre-taining. Furthermore, we use learnable upsampling unit (deconvolutional layers) to reconstruct back the encoded input.

\paragraph{Learnable Upsampling Layer.}The concept of upsampling layer was first introduced by Zeiler and Fergus ~\cite{zeiler14} to visualize the filters of internal layer in a 2D ConvNet.  However, they simply transpose the weights and do not learn the  filter for upsampling. Instead, Long \emph{et al.} ~\cite{Long2015} and Dosovitskiy \emph{et al.}~\cite{DB15}  first introduced the idea of deconvolution (up-sampling) as a trainable layer although for different applications. Recently, using up-sampling to produce spatial output \cite{rematas16cvpr,rematas16arxiv,deeprenderer} - has also seen first applications for computer graphics.  Note that a few concurrent works~\cite{YMitra16,3DRNN-Savarese} also propose a decoder based on volumetric upsampling that outputs 3D reconstruction. In particular, Yumer \emph{et al.}~\cite{YMitra16} uses a similar architecture for predicting deformed version of the input shape. In contrast, we propose a denoising volumetric auto encoder for  shape classification and completion that learns an embedding of shapes in an unsupervised manner.

Rest of the paper is organized as follows: In the next section, we first formulate the problem and describe our deep network and training details. We then move on to the experiment section where we first show the experiments for classification on ModelNet database and the qualitative results for shape interpolation on the learned embedding. We then formulate the protocol for evaluating current techniques for the task of shape completion and show our quantitative results. We conclude with qualitative results for shape completion and denoising.

\section{Unsupervised Learning of Volumetric Representation by Completion}
Given a collection of shapes of various objects and their different poses, our aim is to learn the shape distributions of various classes by predicting the missing voxels from the rest. Later, we want to leverage the learnt embedding for shape recognition and interpolation tasks as well as use the generative capabilities of the auto encoder architectures for predicting enhanced version of  corrupted representations. These corruptions can range from noise like missing voxels to more severe structured noise patterns.

\subsection{VConv-DAE: Fully Convolutional Denoising Auto Encoder }

\paragraph{Voxel Grid Representation.} Following Shapenet\cite{Shape_net}, we adopt the same input representation of a geometric shape: a voxel cube of resolution $24^3$. Thereafter, each mesh is first converted to a voxel representation with 3 extra cells of padding in both directions to reduce the convolution border artifacts and stored as binary tensor where 1 indicates the voxel is inside the mesh surface and 0 indicates the voxel is outside the mesh. This results in the overall dimensions of voxel cube of size 30$\times$30$\times$30.

\begin{figure}[t]
        \centering
        \includegraphics[width=0.9\linewidth]{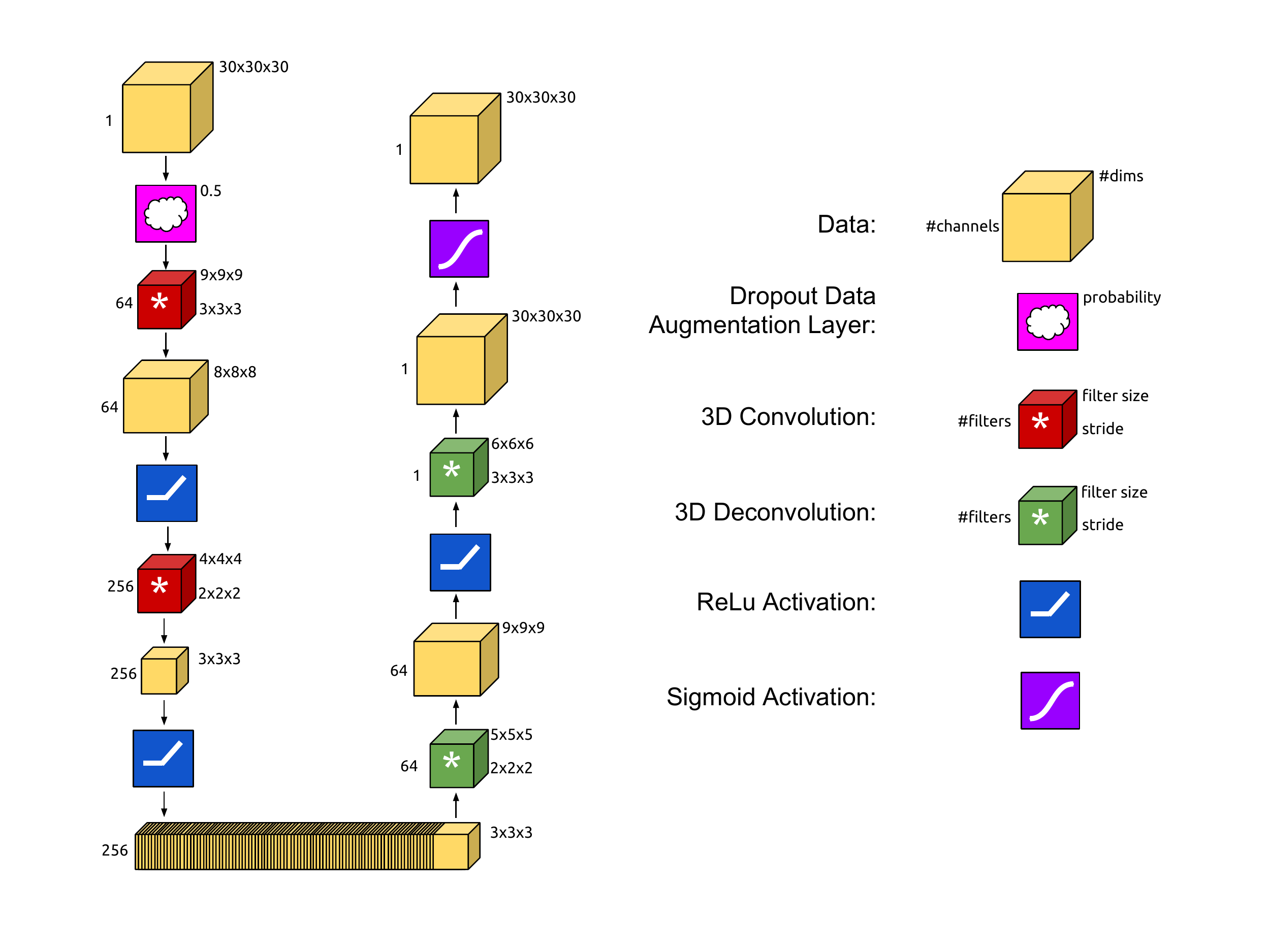}
        \caption{VConv-DAE: Convolutional Denoising Auto Encoder for Volumetric Representations}\label{fig:dae}

\end{figure}

\paragraph{Overview of Architecture.}
To this end, we learn an end to end, voxel to voxel mapping by phrasing it as two class (1-0) auto encoder formulation from  a whole voxel grid to a whole voxel grid.
An overview of our VConv-DAE architecture is shown in Figure \ref{fig:dae}.
Labels in our training corresponds to the voxel occupancy and not class label. Our architecture starts with a dropout layer directly connected to the input layer. The left half of our network can be seen as an encoder stage that results in a condensed representation (bottom of figure) which is connected to a fully connected layer in between. In the second stage (right half), the network reconstructs back the input from this intermediate representation by deconvolutional(Deconv) layers which acts as a learnable local up-sampling unit. We will now explain the key components of the architecture in more detail.

\paragraph{Data Augmentation Layer.} While data augmentation has been used a lot to build deep invariant features for images ~\cite{krizhevsky2012imagenet}, it is relatively little explored on volumetric data. We put a dropout~\cite{JMLR:v15:srivastava14a} layer on the input. This serves the purpose of input data augmentation and an implicit training on a virtually infinite amount of data and has shown in our experiments to greatly avoid over-fitting.

 \paragraph{Encoding Layers: 3D Convolutions }
 The first convolutional layer has 64 filters of size 9 and stride 3. The second convolutional layer has 256 filters of size 4 and stride 2 meaning each filter has 64$\times$4$\times$4$\times$4 parameters. This results into 256 channels of size 3$\times$3$\times$3. These feature maps are later  flattened   into one dimensional vector of length 6912 (= 256$\times$3$\times$3$\times$3) which is followed by a fully connected layer of same length (6912).  This bottleneck layer later acts as a shape embedding for classification and interpolation experiments later. The fixed size encoded input is now reconstructed back with two deconv layers. First Deconv layer contains 64 filters of size 5 and stride 2 while the last deconv layer finally merges all 64 feature cubes back to the original voxel grid. It contains a filter of size 6 and stride 3.

\paragraph{Decoding Layers: 3D Deconvolutions} While CNN architecture based on convolution operator have been very powerful and effective in a range of vision problems, Deconvolutional (also called convolutional transpose) based architecture are gaining traction recently.
 Deconvolution (Deconv) is basically convolution transpose which takes one value from the
input, multiplies the value by the weights in the filter, and
place the result in the output channel. Thus, if the 2D filter
has size f$\times$ f, it generates a f$\times$f output matrix for
each pixel input. The output is generally stored with a
overlap (stride ) in the output channel. Thus, for input x, filter size f,
and stride d, the output is of dims $(x-i)*d +f$. Upsampling is performed until the original size of the input has been regained.

We did not extensively experiment with different network configurations. However, small variations in network depth and width did not seem to have significant effect on the performance. Some of the design choice  also take into account the input voxel resolution. We chose two convolutional layers for encoder to extract robust features at multiple scales. Learning a robust shape representation essentially means capturing the correlation between different voxels. Thus, receptive field of convolutional filter plays a major role and we observe the best performance with large conv filters in the first layer and large deconv filter in the last layer. We experimented with two types of loss functions: mean square loss and cross entropy loss. Since we have only two classes, there is not much variation in performance with cross entropy being slightly better.

\subsection{Dataset and Training}
\paragraph{Dataset.}
 Wu et al.~\cite{Shape_net} use Modelnet, a large scale 3D CAD model dataset for their experiments. It contains 151,128 3D CAD models belonging to 660 unique object categories. They provide two subset of this large scale dataset for the experiments. The first subset contains 10 classes that overlaps with the NYU dataset~\cite{Silberman} and  contains  indoor scene classes such as sofa, table, chair, bed etc.  The second subset of the dataset contains 40 classes where each class has at least 100 unique CAD models.
Following the protocol of ~\cite{Shape_net}, we use both 10 and 40 subset for classification while completion is restricted to subset of 10 that mostly corresponds to indoor scene objects.
\paragraph{Training Details.}
We train our network end-to-end from scratch. We experiment with different levels of dropout noise and observe that training with more noisy data helps in generalising well for the task of denoising and shape completion. Thus, we set a noise level of $p= 0.5$ for our Dropout data augmentation layer which eliminates half of the input at random and therefore the network is trained for reconstruction by only observing $50\%$ of the input voxels. We train our network with pure stochastic gradient descent and a learning rate of $0.1$ for $500$ epochs. We use momentum with a value of $0.9$. We use the open source library Torch for implementing our network and will make our code public at the time of publication.

\section{Experiments}
We conduct a series of experiments in particular to establish a comparison to the related work of Shapenet \cite{Shape_net}. First, we evaluate the representation that our approach acquires in an unsupervised way on a classification task and thereby directly comparing to Shapenet. We then support the empirical performance of feature learning with qualitative results obtained by linear interpolation on the embedding space of various shapes. Thereafter, we propose two settings to evaluate quantitatively the generative performance of 3D deep learning approach on a denoising and shape completion task -- on which we also benchmark against Shapenet and baselines related to our own setup. We conclude the experiments with qualitative results.

\subsection{Evaluating the Unsupervised Embedding Space}
Features learned from deep networks are state-of-the-art in various computer vision problems. However, unsupervised volumetric feature learning is a less explored area. Our architecture is primarily designed for shape completion and denoising task. However, we are also interested in evaluating how the features learned in unsupervised manner compare with fully supervised state-of-the-art 3D classification only architecture.

\subsubsection{Classification Setup.}
We conduct 3D classification  experiments to evaluate our features. Following shapenet, we use the same train/test split by taking the first 80 models for training and first 20 examples for test. Each CAD model is rotated along gravity direction every  30 degree which results in total 38,400 CAD models for training and 9,600 for testing.
We propose following two methods to evaluate our network for the task of classification:
\begin{enumerate}
\item Ours-UnSup : We feed forward the test set to the network and simply take the fixed length bottleneck layer of dimensions 6912 and use this as a feature vector for a linear SVM. Note that the representation is trained completely unsupervised.
\item Ours-Fine Tuned(FT): We follow the set up of Shapenet~\cite{Shape_net} which puts a layer with class labels on the top most feature layer and fine tunes the network. In our network, we take the bottleneck layer which is of 6912 dimensions and  put another layer in between bottleneck and softmax layer. So, the resulting classifier has an intermediate fully connected layer 6912-512-40.\\
\end{enumerate}
For comparison, we also report performance for Light Field descriptor (LFD~\cite{chen2003visual}, 4,700 dimensions)
and Spherical Harmonic descriptor (SPH~\cite{kazhdan2003rotation}, 544 dimensions). We also report the overall best performance achieved so far on this dataset~\cite{su15mvcnn,Maturana_2015_7900}.

\begin{table}[h]
\centering
\begin{tabular}{ c |c c c c  c|c c }
 \hline
 10 classes & SPH~\cite{kazhdan2003rotation} & LFD~\cite{chen2003visual} & SN~\cite{Shape_net} & Ours-UnSup & Ours-FT & VoxNet & MvCnn~\cite{su15mvcnn} \\
 \hline

  AP   &  79.79 &  79.87  & 83.54 &  80.50  & \textbf{84.14} & \textbf{92}  &   \\
 \hline
 40 classes & SPH~\cite{kazhdan2003rotation} & LFD~\cite{chen2003visual} & SN~\cite{Shape_net} & Ours-UnSup & Ours-FT  & VoxNet & MvCnn~\cite{su15mvcnn} \\
 \hline

  AP   &  68.23  & 75.47   &  77.32 & 75.50 & \textbf{79.84}  &  83.00 & \textbf{90.10}  \\
 \hline
\end{tabular}
\caption{\small Shape Classification Results }
\end{table}

\subsubsection{Discussion}
Our representation, Ours-UnSup, achieves 75 \% accuracy on the 40 classes while trained completely unsupervised. When compared to the setup of fine tuned shapenet, our fine tuned representation, Ours-FT, compares favorably and outperforms shapenet on both 10 and 40 classes.

\paragraph{Comparison with the state-of-the-art.} Our architecture is designed for shape completion and denoising while Voxnet and MvCnn is for recognition only. We demonstrate that it also lends to recognition and shows promising performance. MvCnn ~\cite{su15mvcnn} outperforms other methods by a large margin. However, unlike the rest of the methods, it works in the image domain. For each model, MvCnn ~\cite{su15mvcnn} first renders it in different views and then aggregates the scores of all rendered images obtained by CNN. Compared to the fully supervised classification network of Voxnet, our accuracy is only 3 \%  less on 40 classes. This is because Voxnet architecture is shallow and contains pooling layers that help classification but are not suitable for reconstruction purpose, as also noted in 3D Shapenet[4].
\subsection{Linear Interpolation in the embedding space.} 

\newlength{\imagewidth}
\settowidth{\imagewidth}{\includegraphics{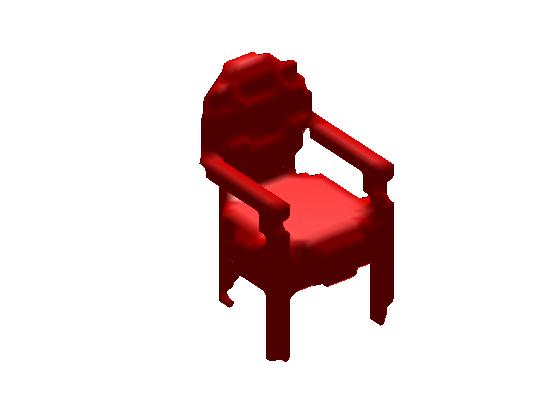}}
\begin{table}[h!]
\begin{center}
\begin{tabular}{c|cccc|c}  \\
Source(t=1)& t= 3 & t=5 & t=7  & t = 9 &Target(t= 10) \\\hline

&&&&&\\
 \includegraphics[trim=0.2\imagewidth{} 0.1\imagewidth{} 0.2\imagewidth{} 0.05\imagewidth{}, clip, width = 0.085\imagewidth{}]{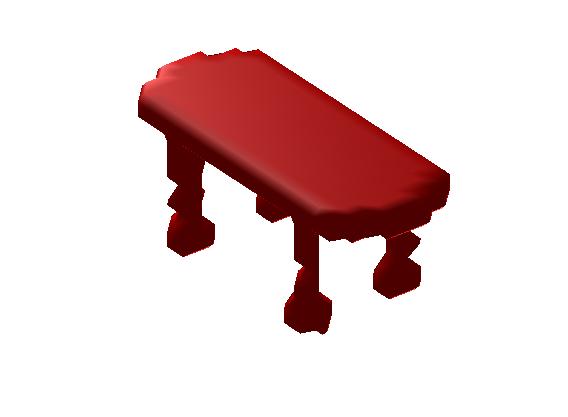} &
\includegraphics[trim=0.2\imagewidth{} 0.1\imagewidth{} 0.2\imagewidth{} 0.05\imagewidth{}, clip, width = 0.085\imagewidth{}]{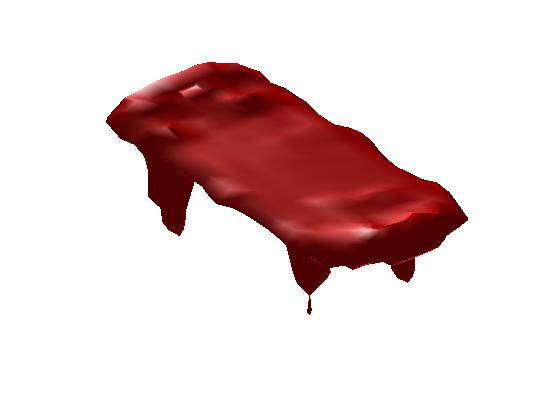} &
\includegraphics[trim=0.2\imagewidth{} 0.1\imagewidth{} 0.1\imagewidth{} 0.05\imagewidth{}, clip, width = 0.085\imagewidth{}]{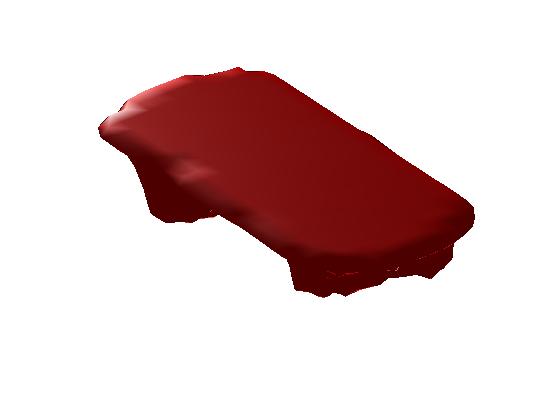} &
\includegraphics[trim=0.2\imagewidth{} 0.1\imagewidth{} 0.1\imagewidth{} 0.05\imagewidth{}, clip, width = 0.085\imagewidth{}]{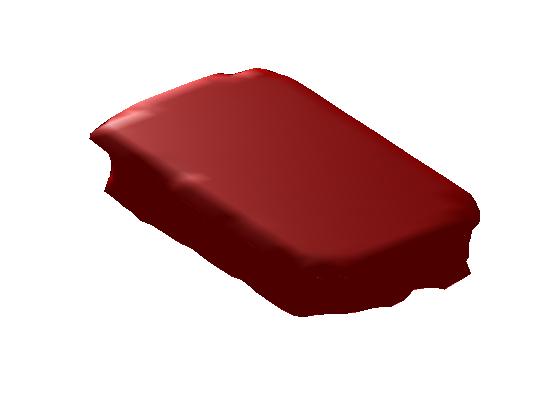} &
\includegraphics[trim=0.2\imagewidth{} 0.1\imagewidth{} 0.1\imagewidth{} 0.05\imagewidth{}, clip, width = 0.085\imagewidth{}]{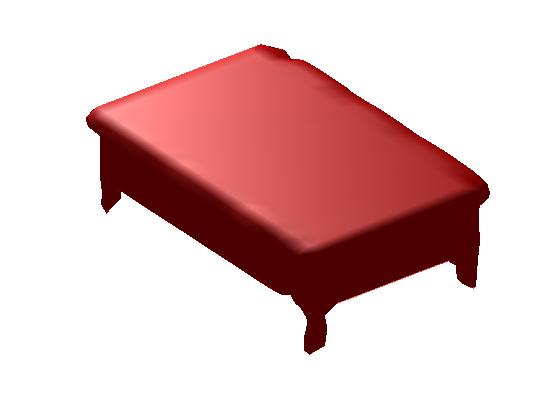} &
\includegraphics[trim=0.2\imagewidth{} 0.1\imagewidth{} 0.1\imagewidth{} 0.05\imagewidth{}, clip, width = 0.085\imagewidth{}]{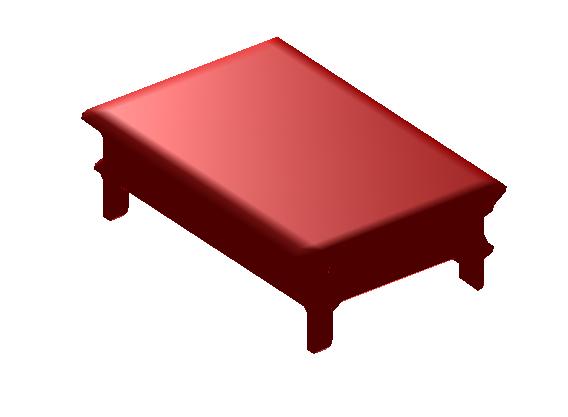}
\\
\includegraphics[trim=0.2\imagewidth{} 0.1\imagewidth{} 0.2\imagewidth{} 0.05\imagewidth{}, clip, width = 0.085\imagewidth{}]{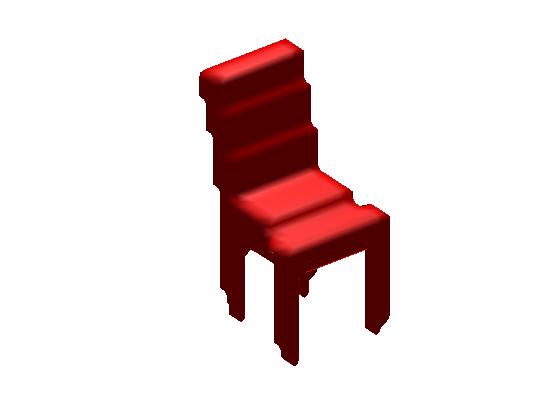} &
\includegraphics[trim=0.2\imagewidth{} 0.1\imagewidth{} 0.2\imagewidth{} 0.05\imagewidth{}, clip, width = 0.085\imagewidth{}]{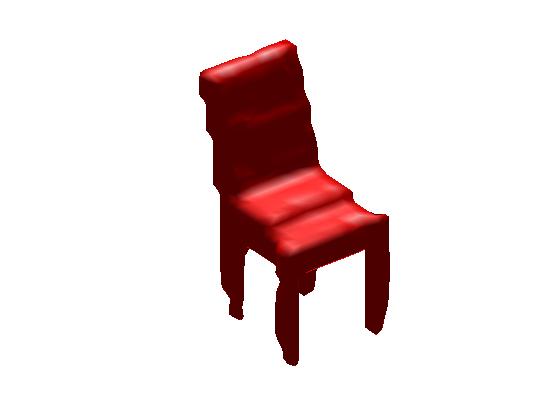} &
\includegraphics[trim=0.2\imagewidth{} 0.1\imagewidth{} 0.2\imagewidth{} 0.05\imagewidth{}, clip, width = 0.085\imagewidth{}]{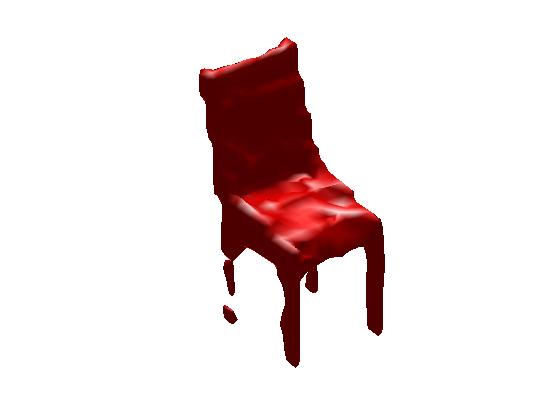} &
\includegraphics[trim=0.2\imagewidth{} 0.1\imagewidth{} 0.2\imagewidth{} 0.05\imagewidth{}, clip, width = 0.085\imagewidth{}]{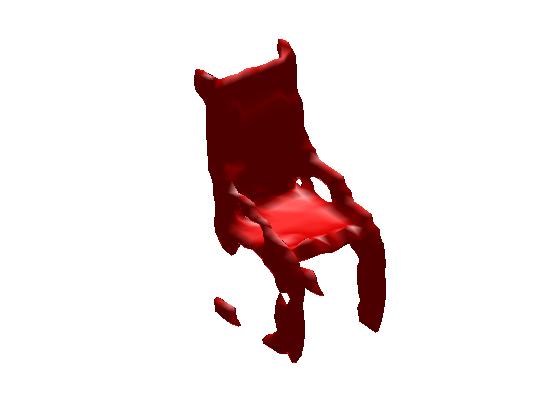}  &
\includegraphics[trim=0.2\imagewidth{} 0.1\imagewidth{} 0.2\imagewidth{} 0.05\imagewidth{}, clip, width = 0.085\imagewidth{}]{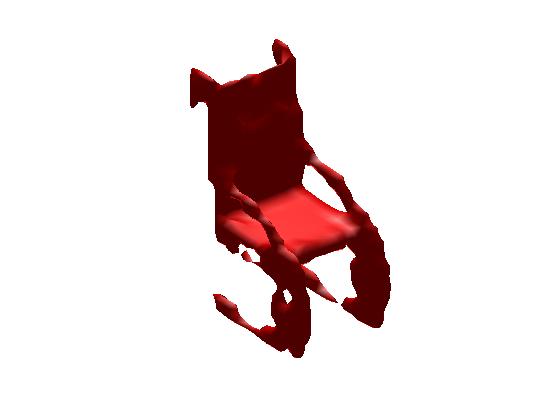}  &
\includegraphics[trim=0.2\imagewidth{} 0.1\imagewidth{} 0.2\imagewidth{} 0.05\imagewidth{}, clip, width = 0.085\imagewidth{}]{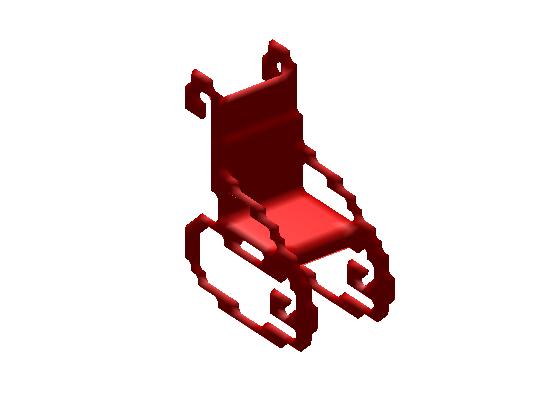}
\\
\includegraphics[trim=0.2\imagewidth{} 0.1\imagewidth{} 0.1\imagewidth{} 0.05\imagewidth{}, clip, width = 0.085\imagewidth{}]{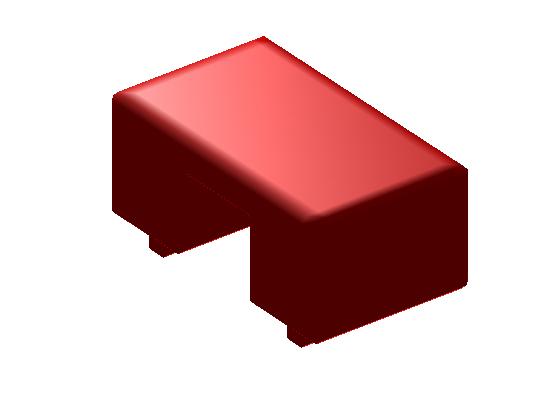}&
\includegraphics[trim=0.2\imagewidth{} 0.1\imagewidth{} 0.1\imagewidth{} 0.05\imagewidth{}, clip, width = 0.085\imagewidth{}]{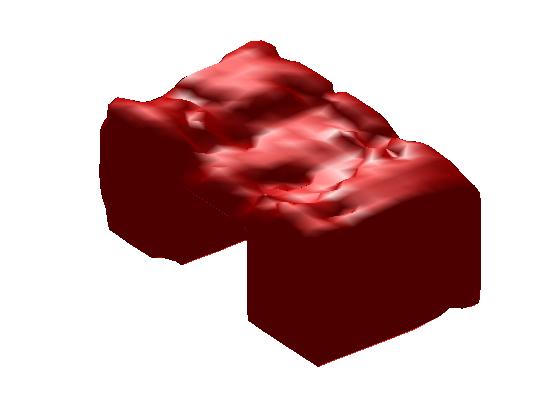}&
\includegraphics[trim=0.2\imagewidth{} 0.1\imagewidth{} 0.1\imagewidth{} 0.05\imagewidth{}, clip, width = 0.085\imagewidth{}]{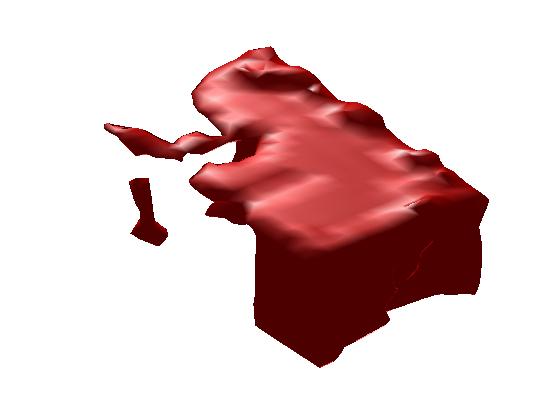}&
\includegraphics[trim=0.2\imagewidth{} 0.1\imagewidth{} 0.1\imagewidth{} 0.05\imagewidth{}, clip, width = 0.085\imagewidth{}]{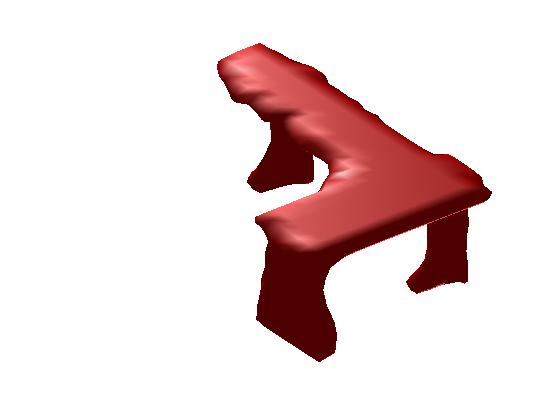}&
\includegraphics[trim=0.2\imagewidth{} 0.1\imagewidth{} 0.1\imagewidth{} 0.05\imagewidth{}, clip, width = 0.085\imagewidth{}]{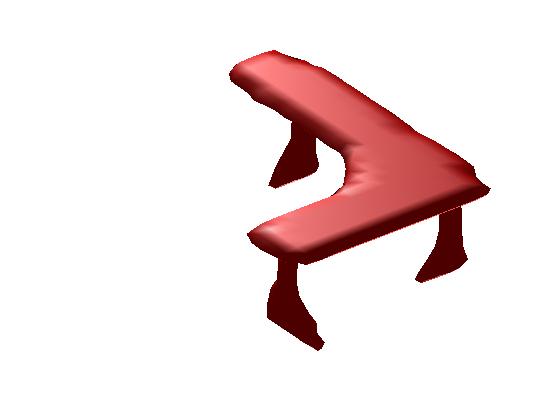}&
\includegraphics[trim=0.2\imagewidth{} 0.1\imagewidth{} 0.1\imagewidth{} 0.05\imagewidth{}, clip, width = 0.085\imagewidth{}]{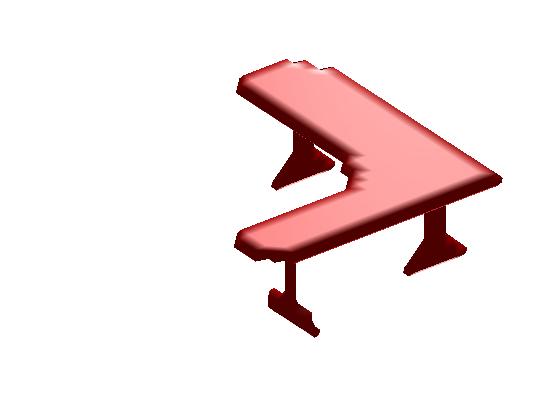}\\

\includegraphics[trim=0.2\imagewidth{} 0.1\imagewidth{} 0.1\imagewidth{} 0.05\imagewidth{}, clip, width = 0.085\imagewidth{}]{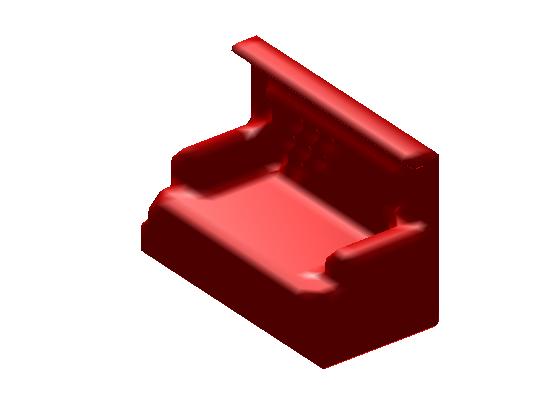}&
\includegraphics[trim=0.2\imagewidth{} 0.1\imagewidth{} 0.1\imagewidth{} 0.05\imagewidth{}, clip, width = 0.085\imagewidth{}]{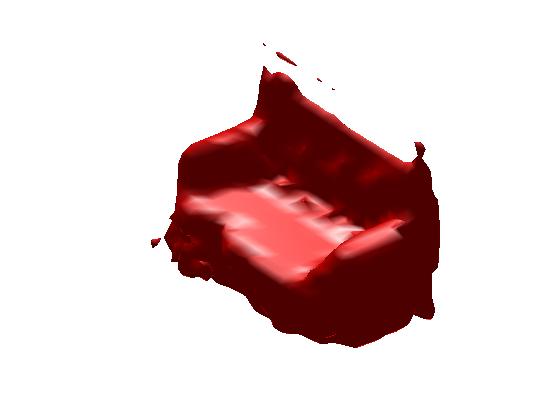}&
\includegraphics[trim=0.2\imagewidth{} 0.1\imagewidth{} 0.1\imagewidth{} 0.05\imagewidth{}, clip, width = 0.085\imagewidth{}]{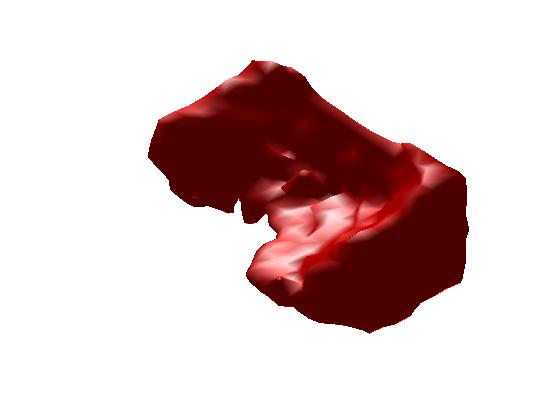}&
\includegraphics[trim=0.2\imagewidth{} 0.1\imagewidth{} 0.1\imagewidth{} 0.05\imagewidth{}, clip, width = 0.085\imagewidth{}]{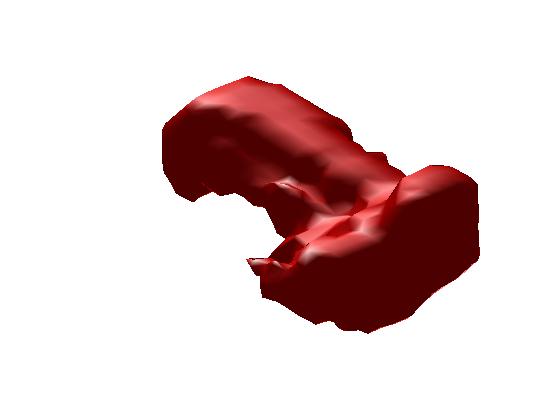}&
\includegraphics[trim=0.2\imagewidth{} 0.1\imagewidth{} 0.1\imagewidth{} 0.05\imagewidth{}, clip, width = 0.085\imagewidth{}]{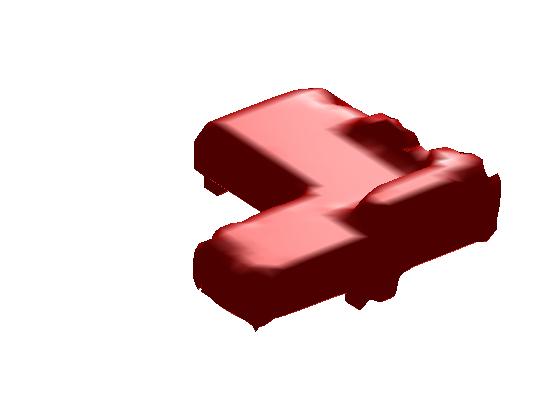}&
\includegraphics[trim=0.2\imagewidth{} 0.1\imagewidth{} 0.1\imagewidth{} 0.05\imagewidth{}, clip, width = 0.085\imagewidth{}]{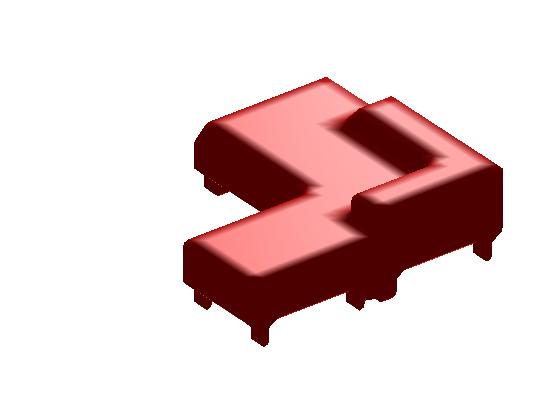}\\
\includegraphics[trim=0.2\imagewidth{} 0.1\imagewidth{} 0.1\imagewidth{} 0.05\imagewidth{}, clip, width = 0.085\imagewidth{}]{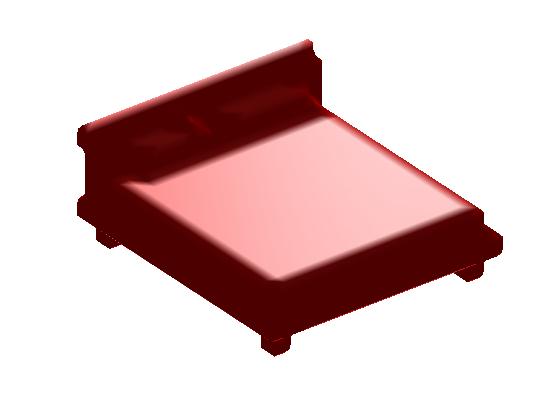}&
\includegraphics[trim=0.2\imagewidth{} 0.1\imagewidth{} 0.1\imagewidth{} 0.05\imagewidth{}, clip, width = 0.085\imagewidth{}]{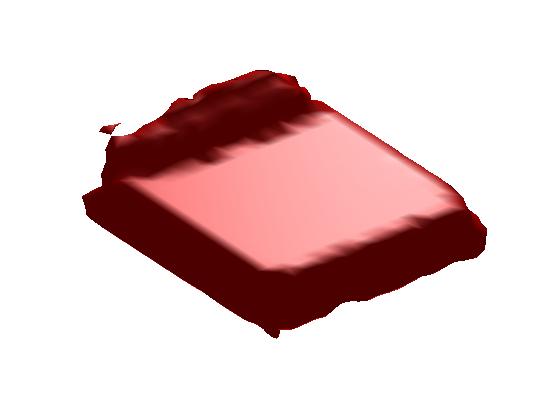}&
\includegraphics[trim=0.2\imagewidth{} 0.1\imagewidth{} 0.1\imagewidth{} 0.05\imagewidth{}, clip, width = 0.085\imagewidth{}]{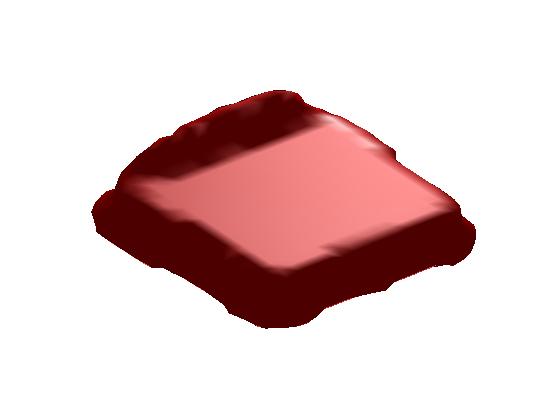}&
\includegraphics[trim=0.2\imagewidth{} 0.1\imagewidth{} 0.1\imagewidth{} 0.05\imagewidth{}, clip, width = 0.085\imagewidth{}]{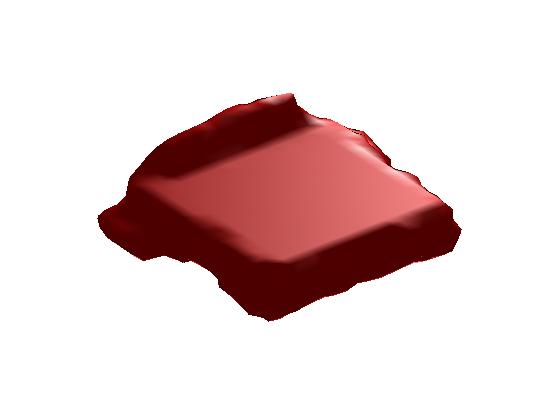}&
\includegraphics[trim=0.2\imagewidth{} 0.1\imagewidth{} 0.1\imagewidth{} 0.05\imagewidth{}, clip, width = 0.085\imagewidth{}]{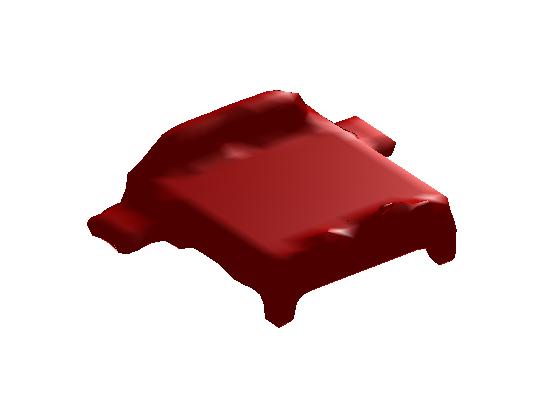}&
\includegraphics[trim=0.2\imagewidth{} 0.1\imagewidth{} 0.1\imagewidth{} 0.05\imagewidth{}, clip, width = 0.085\imagewidth{}]{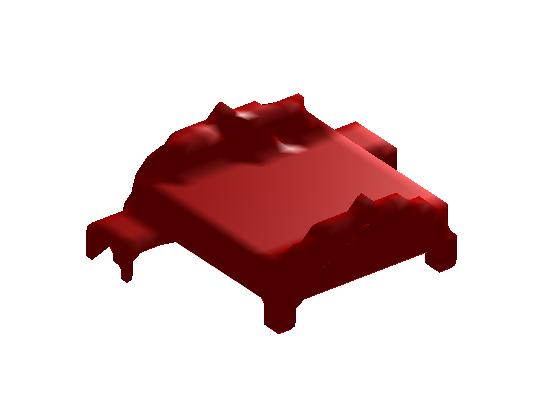}
\end{tabular}
\end{center}
\caption{Shape Interpolation results}\label{fig:shape-inter}
\end{table}
Encouraged by the unsupervised volumetric feature learning performance, we analyze the representation further to understand the embedding space learned by our auto-encoder. To this end, we randomly choose two different instances of a class in the same pose as input and then, feed it to the encoder part of our trained model which transforms any shape into a fixed length encoding vector of length 6912. We call these two instances Source and Target in Table \ref{fig:shape-inter}. On a scale from 1 to 10, where 1 corresponds to source instance and 10 to target, we then linearly interpolate the eight intermediate encoded vectors. These interpolated vectors are then fed to the second (decoder) part of the model which decodes the encoded vector into volumes. Note that we chose linear interpolation over non-linear partially because of the simplicity and the fact that the feature space already achieves linear separability on 10 classes with an accuracy of 80 \%. In Table \ref{fig:shape-inter}, we show the interpolated volumes between source and target at each alternative step. We observe that in most cases new, connected shapes are inferred as intermediate steps and a plausible transition is produced even for highly non-convex shapes.
\section{Denoising and Shape Completion}
In this section, we show experiments that evaluate our network on the task of denoising and shape completion.  This is very relevant in the scenario where geometry is captured with depth sensors that often comes with holes and noise due to sensor and surface properties. Ideally, this task should be evaluated on real world depth or volumetric data but obtaining ground truth for real world data is very challenging and to the best of our knowledge, there exists no dataset that contains the ground truth for missing parts and holes of Kinect data. Kinect fusion type approach still suffer from sensor failure modes and large objects like furnitures often cannot be scanned from all sides in their typical location.

We thus rely on CAD model dataset where complete geometry of various objects is available and we simulate different noise characteristics to test our network. We use the 10-class subset of ModelNet database for experiments. In order to remain comparable to Shapenet,  we use their pretrained generative model for comparison and train our model on the first 80 (before rotation) CAD models for each class accordingly. This results in 9600 training models for the following experiments.

\subsection{Denoising Experiments} We first evaluate our network on the same random noise characteristics with which we train the network. This is  challenging since the test set contains different instances than training set. We also vary the amount of random noise injected during test time for evaluation.  Training is same as that for classification and we use the same model trained on the first 80 CAD models. At test time, we use all the test models available for these 10 classes for evaluation.

\paragraph{Baseline and Metric}
To better understand the network performance for reconstruction at test time, we study following methods:
\begin{enumerate}
\item Convolutional Auto Encoder (CAE): We train the same network without any noise meaning without any dropout layer. This baseline tells us the importance of data augmentation or injecting noise during training.

\item Shapenet(SN): We use pretrained generative model for 10 class subset made public by Shapenet and their code for completion. We use the same hyper-parameters as given in the their source code for completion. Therefore, we set number of epochs to $50$, number of Gibbs iteration to 1 and threshold parameter to $0.1$. Their method assumes that an object mask is available for the task of completion at test time. Our model does not  make  such assumption since this is difficult to obtain at test time.
Thus, we evaluate shapenet with two different scenario for such a mask: first, SN-1,  by setting the whole voxel grid as mask and second, SN-2, by  setting the occupied voxels in test input as mask. Given the range of hyper-parameters, we report performance for the best hyperparameters.

\end{enumerate}
 \begin{table}[h]
\centering
\begin{tabular}{ p{2cm}|p{1cm} p{1cm} p{1cm} p{1cm}|p{1cm} p{1cm} p{1cm} p{1cm}}
 \hline
\multirow{2}{*}{Class} &
      \multicolumn{4}{c|}{30\% noise} &
      \multicolumn{4}{c}{50\% noise} \\

 &  CAE & SN-2 & SN-1& Ours & CAE & SN-2 & SN-1& Ours  \\
 \hline
 \hline
 Bed     &   2.88   & \hspace{0.15cm}6.88 & \hspace{0.15cm}6.76   & \textbf{0.83} & \hspace{0.15cm}6.56 & \hspace{0.15cm}9.87& \hspace{0.15cm}7.25 & \textbf{1.68}  \\
 Sofa      & 2.60 & \hspace{0.15cm}7.48 &   \hspace{0.15cm}7.97 &\textbf{0.74}  & \hspace{0.15cm}6.09  & \hspace{0.15cm}9.51 & \hspace{0.15cm}8.67 &\textbf{1.81} \\
 Chair     &   2.51 & \hspace{0.15cm}6.73  & 11.76  &\textbf{1.62}  &   \hspace{0.15cm}4.98  & \hspace{0.15cm}7.82& 11.97 &\textbf{2.45} \\
 Desk      &   2.26 & \hspace{0.15cm}7.94 &   10.76  &\textbf{1.05} &  \hspace{0.15cm}5.38  & \hspace{0.15cm}9.35 & 11.04 &\textbf{1.99}   \\
 Toilet      &  4.25 & 16.05 &  17.92 &\textbf{1.57} & \hspace{0.15cm}9.94   & 17.95& 18.42 &\textbf{3.36}     \\
 Monitor    &    3.01 & 11.30 &  14.75 &\textbf{1.26} & \hspace{0.15cm}7.01 & 12.42& 14.95 & \textbf{2.37}    \\
 Table      &   1.17 & \hspace{0.15cm}3.47 &  \hspace{0.15cm}5.77  &\textbf{0.53} &   \hspace{0.15cm}2.79 & \hspace{0.15cm}4.65& \hspace{0.15cm}5.88  &\textbf{0.80}  \\
 Night-stand  &  5.07  & 20.00 &  17.90 &\textbf{1.20} & 13.57  & 25.15& 20.49 &\textbf{2.50} \\
 Bathtub    &   2.56  &  \hspace{0.15cm}6.71 & 10.11  &\textbf{0.97} &   \hspace{0.15cm}5.30 &  \hspace{0.15cm}8.26& 10.51   &\textbf{1.77}\\
 Dresser    &   5.56 & 20.07  & 18.00 &\textbf{0.70}  &  15.11 & 27.74& 20.00   &\textbf{1.95} \\
 \hline
 \hline
 Mean Error  &  3.18 & 10.66 & 12.10 &\textbf{1.04} &  \hspace{0.15cm}7.67 &  13.27 & 12.91 &\textbf{2.06} \\
  \hline
\end{tabular}
\caption{\small Average error for denoising }
\end{table}

\paragraph{Metric}: We  count the number of voxels which differs from the actual input. So, we take the absolute difference between the reconstructed version of noisy input and original (no-noise) version. We then normalise reconstruction error by total number of voxels in the grid (13824=24$\times$24$\times$24)). Note that the voxel resolution of 30$\times$30$\times$30 is obtained by padding 3 voxels on each side thus network never sees a input with voxel in those padding. This gives us the resulting reconstruction or denoising error in \%.

\begin{table}[h!]
\begin{center}
\begin{tabular}{ccccc}  \\
&Reference & Noisy & Our & 3DShapeNet \\
&Model     & input & Reconstruction & Reconstruction \\\hline
&&&&\\

Chair&
\includegraphics[trim=0.2\imagewidth{} 0.1\imagewidth{} 0.2\imagewidth{} 0.05\imagewidth{}, clip, width = 0.085\imagewidth{}]{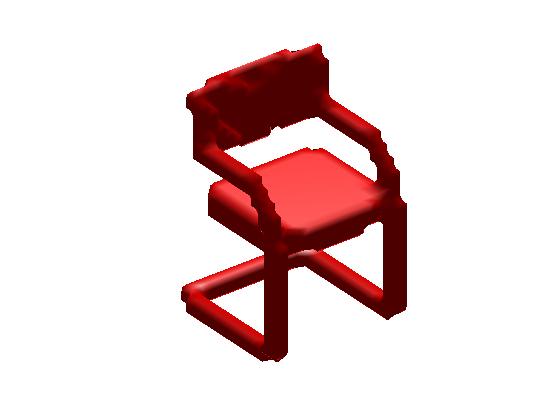} &
\includegraphics[trim=0.2\imagewidth{} 0.1\imagewidth{} 0.2\imagewidth{} 0.05\imagewidth{}, clip, width = 0.085\imagewidth{}]{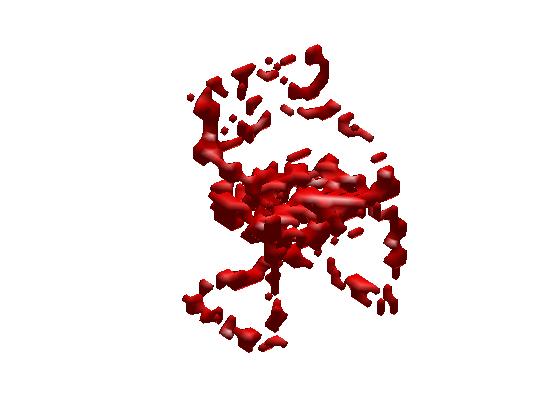} &
\includegraphics[trim=0.2\imagewidth{} 0.1\imagewidth{} 0.2\imagewidth{} 0.05\imagewidth{}, clip, width = 0.085\imagewidth{}]{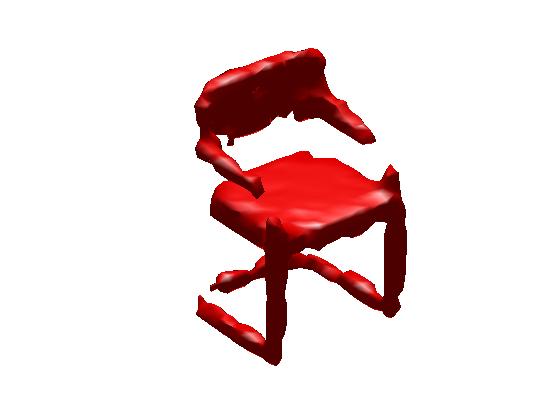} &
\includegraphics[trim=0.2\imagewidth{} 0.1\imagewidth{} 0.2\imagewidth{} 0.05\imagewidth{}, clip, width = 0.085\imagewidth{}]{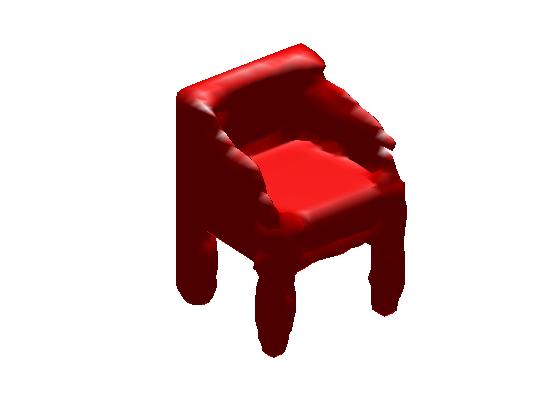} \\

Night Stand &
\includegraphics[trim=0.2\imagewidth{} 0.1\imagewidth{} 0.2\imagewidth{} 0.05\imagewidth{}, clip, width = 0.085\imagewidth{}]{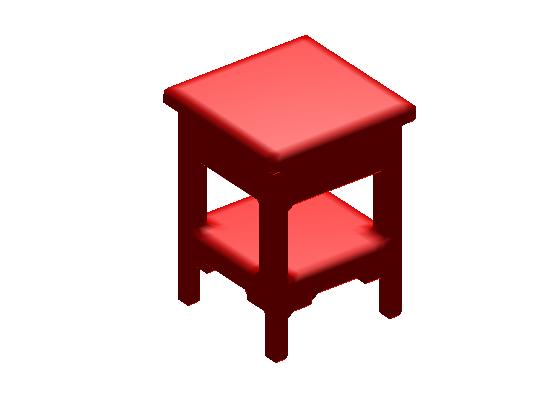} &
\includegraphics[trim=0.2\imagewidth{} 0.1\imagewidth{} 0.2\imagewidth{} 0.05\imagewidth{}, clip, width = 0.085\imagewidth{}]{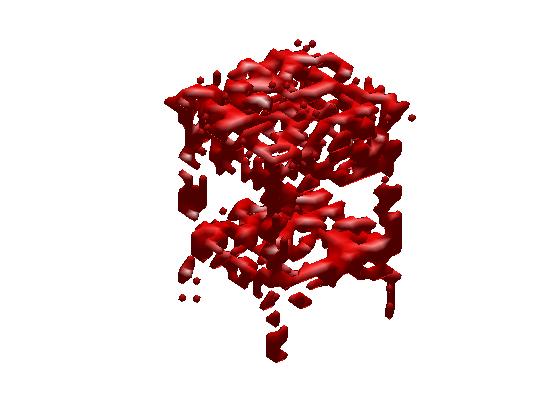} &
\includegraphics[trim=0.2\imagewidth{} 0.1\imagewidth{} 0.2\imagewidth{} 0.05\imagewidth{}, clip, width = 0.085\imagewidth{}]{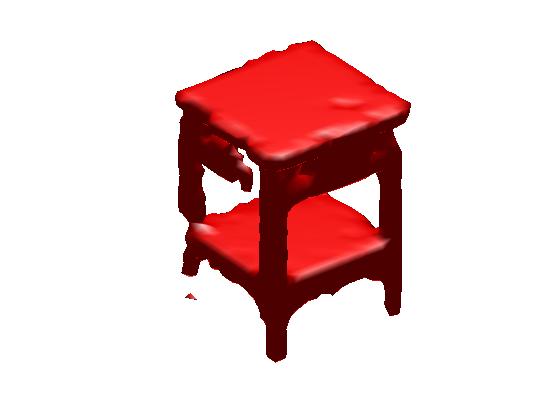} &
\includegraphics[trim=0.2\imagewidth{} 0.1\imagewidth{} 0.2\imagewidth{} 0.05\imagewidth{}, clip, width = 0.085\imagewidth{}]{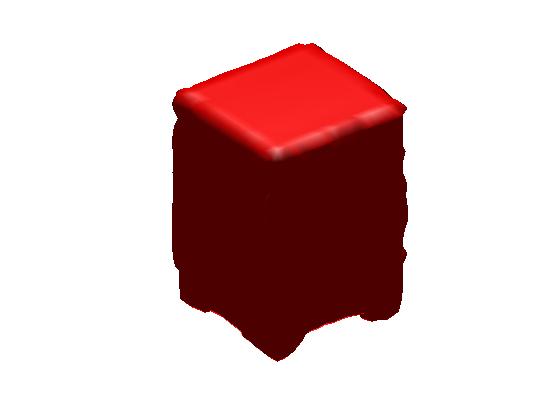} \\

Desk&
\includegraphics[trim=0.2\imagewidth{} 0.1\imagewidth{} 0.1\imagewidth{} 0.05\imagewidth{}, clip, width = 0.085\imagewidth{}]{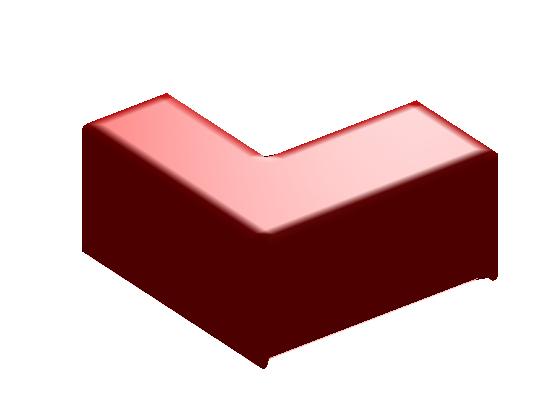} &
\includegraphics[trim=0.2\imagewidth{} 0.1\imagewidth{} 0.1\imagewidth{} 0.05\imagewidth{}, clip, width = 0.085\imagewidth{}]{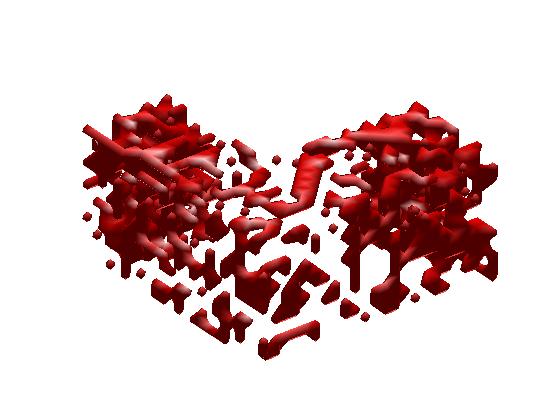} &
\includegraphics[trim=0.2\imagewidth{} 0.1\imagewidth{} 0.1\imagewidth{} 0.05\imagewidth{}, clip, width = 0.085\imagewidth{}]{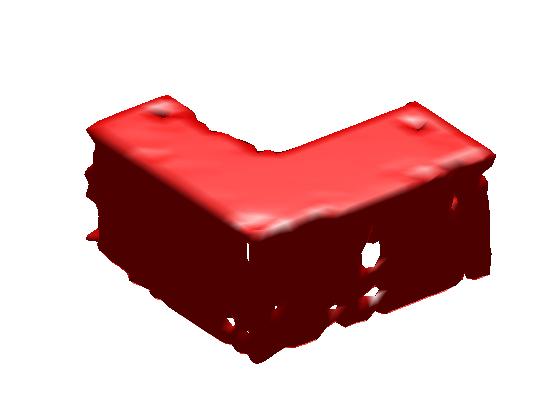} &
\includegraphics[trim=0.2\imagewidth{} 0.1\imagewidth{} 0.1\imagewidth{} 0.05\imagewidth{}, clip, width = 0.085\imagewidth{}]{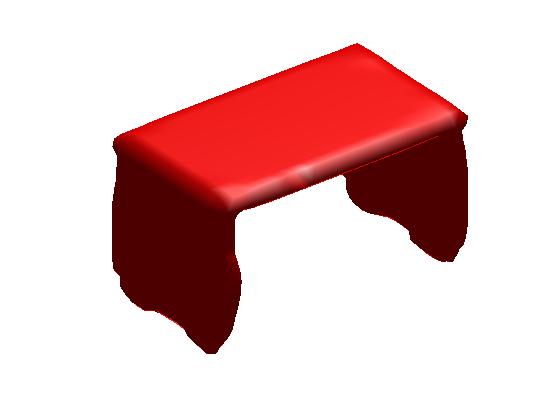} \\

\iffalse
Bathtub &
\includegraphics[trim=0.2\imagewidth{} 0.1\imagewidth{} 0.2\imagewidth{} 0.05\imagewidth{}, clip, width = 0.085\imagewidth{}]{images/bathtub/bathtub_91_orig.jpg} &
\includegraphics[trim=0.2\imagewidth{} 0.1\imagewidth{} 0.2\imagewidth{} 0.05\imagewidth{}, clip, width = 0.085\imagewidth{}]{images/denoising/bathtub_91_50_dist.jpg} &
\includegraphics[trim=0.2\imagewidth{} 0.1\imagewidth{} 0.2\imagewidth{} 0.05\imagewidth{}, clip, width = 0.085\imagewidth{}]{images/denoising/bathtub_91_50_recons.jpg} &
\includegraphics[trim=0.2\imagewidth{} 0.1\imagewidth{} 0.1\imagewidth{} 0.05\imagewidth{}, clip, width = 0.085\imagewidth{}]{images/Shapenet/bathtub_91_SNd.jpg} \\
\fi

Table &
\includegraphics[trim=0.2\imagewidth{} 0.1\imagewidth{} 0.2\imagewidth{} 0.03\imagewidth{}, clip, width = 0.085\imagewidth{}]{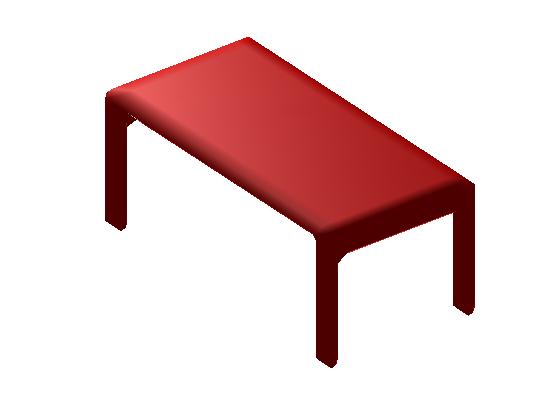} &
\includegraphics[trim=0.2\imagewidth{} 0.1\imagewidth{} 0.2\imagewidth{} 0.03\imagewidth{}, clip, width = 0.085\imagewidth{}]{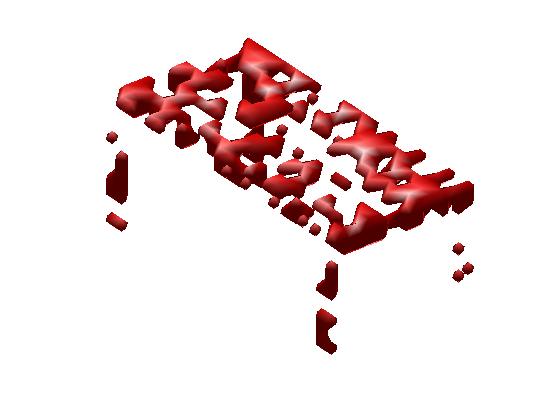} &
\includegraphics[trim=0.2\imagewidth{} 0.1\imagewidth{} 0.2\imagewidth{} 0.03\imagewidth{}, clip, width = 0.085\imagewidth{}]{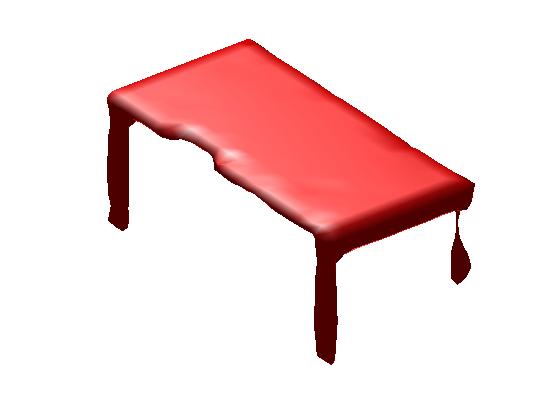} &
\includegraphics[trim=0.2\imagewidth{} 0.1\imagewidth{} 0.1\imagewidth{} 0.03\imagewidth{}, clip, width = 0.085\imagewidth{}]{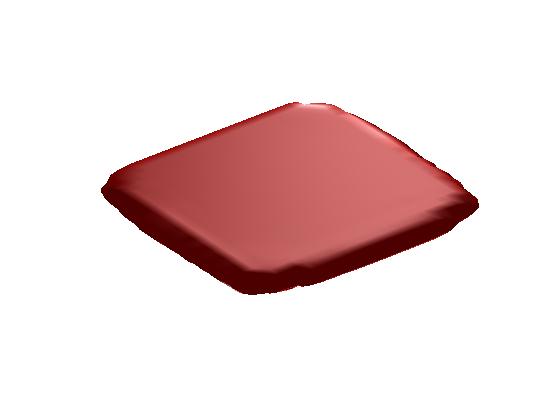}
\\

Sofa &
\includegraphics[trim=0.2\imagewidth{} 0.1\imagewidth{} 0.2\imagewidth{} 0.05\imagewidth{}, clip, width = 0.085\imagewidth{}]{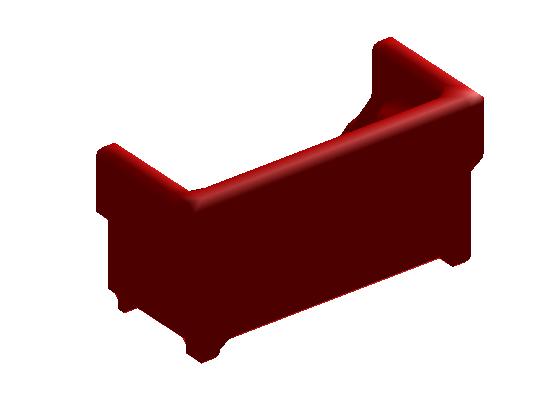} &
\includegraphics[trim=0.2\imagewidth{} 0.1\imagewidth{} 0.2\imagewidth{} 0.05\imagewidth{}, clip, width = 0.085\imagewidth{}]{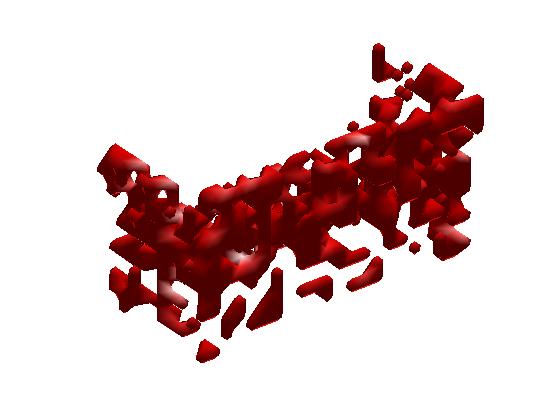} &
\includegraphics[trim=0.2\imagewidth{} 0.1\imagewidth{} 0.2\imagewidth{} 0.05\imagewidth{}, clip, width = 0.085\imagewidth{}]{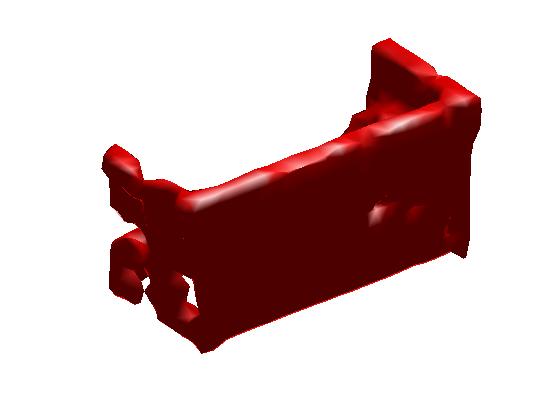} &
\includegraphics[trim=0.2\imagewidth{} 0.1\imagewidth{} 0.1\imagewidth{} 0.05\imagewidth{}, clip, width = 0.085\imagewidth{}]{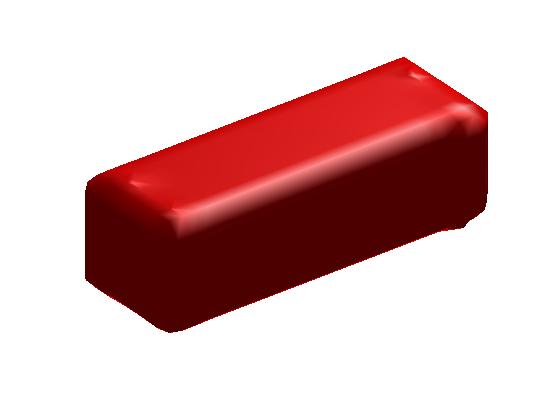} \\
Monitor &
\includegraphics[trim=0.2\imagewidth{} 0.1\imagewidth{} 0.2\imagewidth{} 0.05\imagewidth{}, clip, width = 0.085\imagewidth{}]{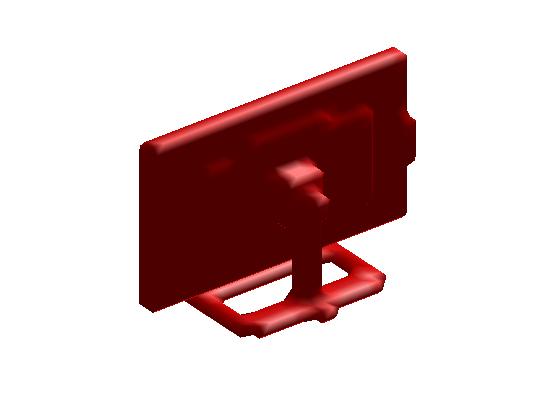} &
\includegraphics[trim=0.2\imagewidth{} 0.1\imagewidth{} 0.2\imagewidth{} 0.05\imagewidth{}, clip, width = 0.085\imagewidth{}]{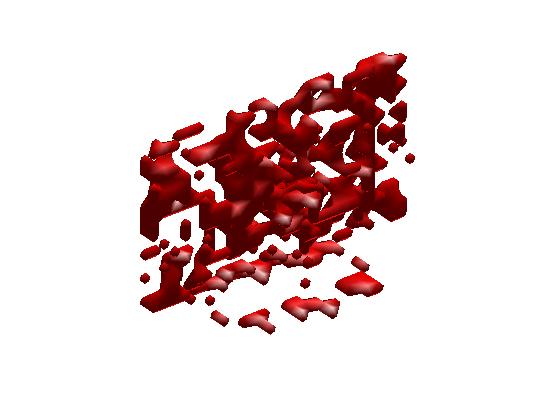} &
\includegraphics[trim=0.2\imagewidth{} 0.1\imagewidth{} 0.2\imagewidth{} 0.05\imagewidth{}, clip, width = 0.085\imagewidth{}]{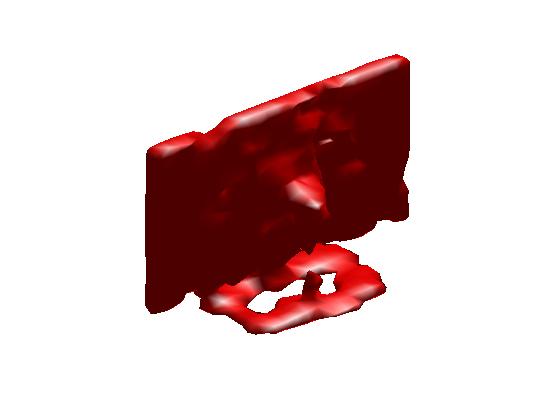} &
\includegraphics[trim=0.2\imagewidth{} 0.1\imagewidth{} 0.2\imagewidth{} 0.05\imagewidth{}, clip, width = 0.085\imagewidth{}]{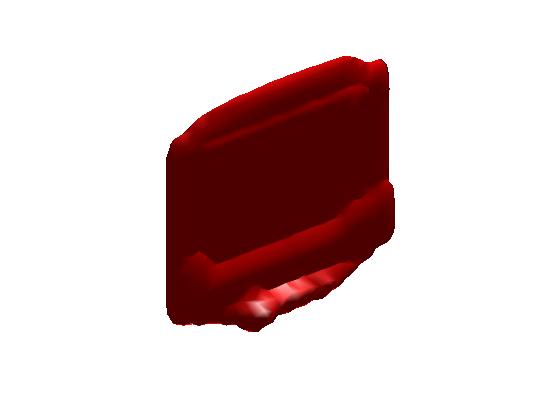} \\

Bed &
\includegraphics[trim=0.2\imagewidth{} 0.1\imagewidth{} 0.05\imagewidth{} 0.05\imagewidth{}, clip, width = 0.085\imagewidth{}]{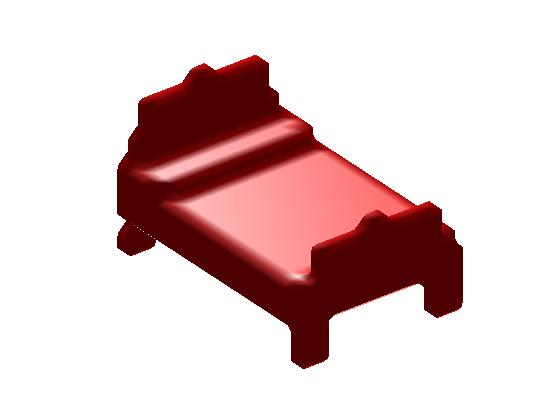} &
\includegraphics[trim=0.2\imagewidth{} 0.1\imagewidth{} 0.05\imagewidth{} 0.05\imagewidth{}, clip, width = 0.085\imagewidth{}]{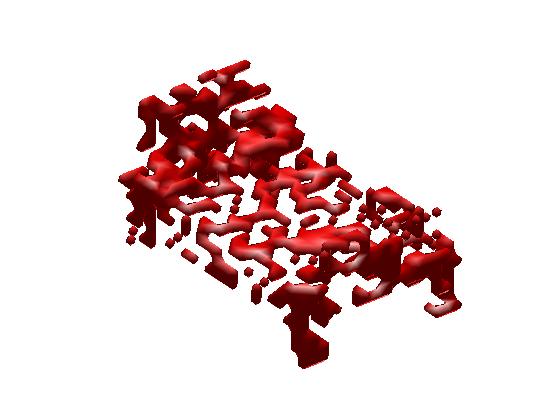} &
\includegraphics[trim=0.2\imagewidth{} 0.1\imagewidth{} 0.05\imagewidth{} 0.05\imagewidth{}, clip, width = 0.085\imagewidth{}]{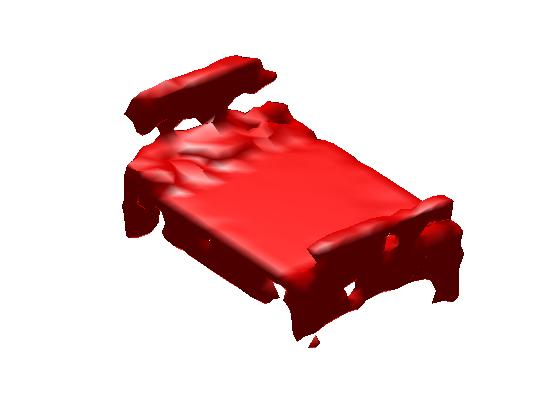} &
\includegraphics[trim=0.2\imagewidth{} 0.1\imagewidth{} 0.05\imagewidth{} 0.05\imagewidth{}, clip, width = 0.085\imagewidth{}]{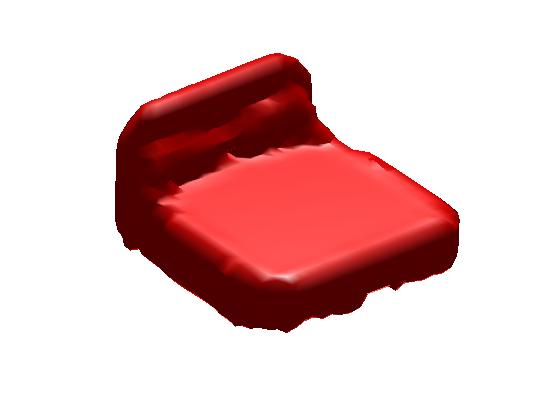} \\

Dresser &
\includegraphics[trim=0.2\imagewidth{} 0.1\imagewidth{} 0.1\imagewidth{} 0.05\imagewidth{}, clip, width = 0.06\imagewidth{}]{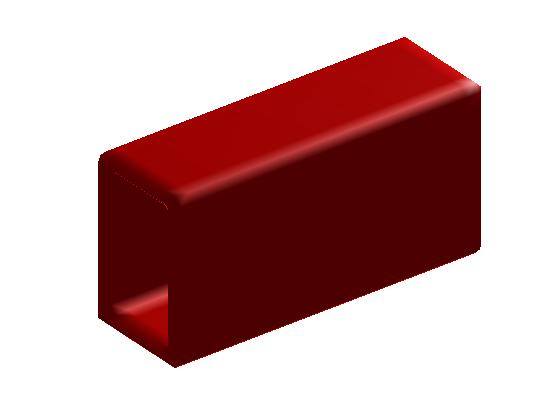} &
\includegraphics[trim=0.2\imagewidth{} 0.1\imagewidth{} 0.1\imagewidth{} 0.05\imagewidth{}, clip, width = 0.06\imagewidth{}]{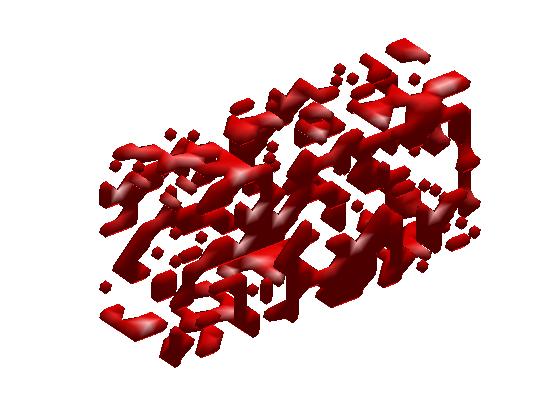} &
\includegraphics[trim=0.2\imagewidth{} 0.1\imagewidth{} 0.1\imagewidth{} 0.05\imagewidth{}, clip, width = 0.06\imagewidth{}]{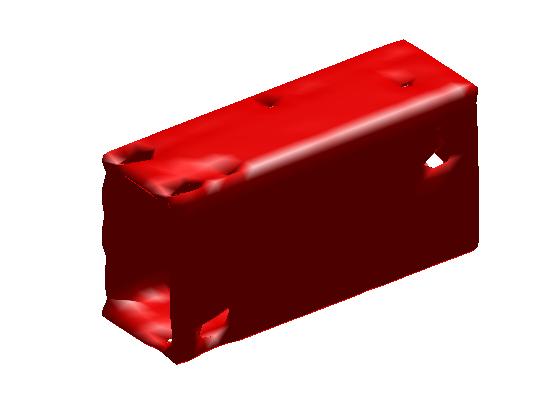} &
\includegraphics[trim=0.2\imagewidth{} 0.1\imagewidth{} 0.05\imagewidth{} 0.05\imagewidth{}, clip, width = 0.06\imagewidth{}]{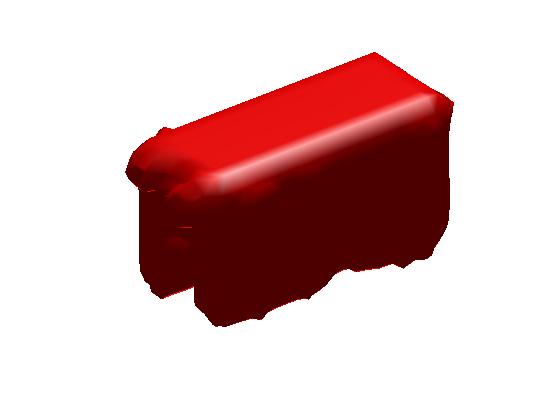} \\
\end{tabular}
\end{center}

\caption{Shape Denoising results for random noise (50\%)}\label{fig:denoising}
\end{table}
\subsection{Slicing Noise and Shape completion}
In this section, we evaluate our network for a structured version of noise that is motivated by occlusions in real world scenario failure modes of the sensor which generates ``holes'' in the data. To simulate such scenarios, we inject slicing noise in the test set as follows: For each instance, we first randomly choose n slices of volumetric cube and remove them. We then evaluate our network on three amount of slicing noise depending upon how many slices are removed.
\begin{table}[h]
\centering
\begin{tabular} { p{2cm}|p{1cm} p{1cm} p{1cm} | p{1cm} p{1cm} p{1cm}| p{1cm} p{1cm} p{1cm}}
 \hline
 \multirow{2}{*}{Class} &
      \multicolumn{3}{c}{10\%} &
      \multicolumn{3}{c}{20\%} &
      \multicolumn{3}{c}{30\%} \\
    & SN-1  &SN-2 & Ours &SN-1 & SN-2 &  Ours & SN-1  & SN-2 & Ours  \\
 \hline
 Bed   & \hspace{0.15cm}7.09 & \hspace{0.15cm}4.71  &  \textbf{1.11} & \hspace{0.15cm}7.25  & \hspace{0.15cm}5.70  & \textbf{1.63}  & \hspace{0.15cm}7.53 &\hspace{0.15cm}6.89 & \textbf{2.40}    \\
 Sofa  &   \hspace{0.15cm}8.05&\hspace{0.15cm}5.51  &  \textbf{0.97} & \hspace{0.15cm}8.32 & \hspace{0.15cm}6.39  & \textbf{1.51}& \hspace{0.15cm}8.66  &\hspace{0.15cm}7.23 & \textbf{2.33}    \\
 Chair  & 12.22 & \hspace{0.15cm}5.66  & \textbf{1.64}& 12.22  & \hspace{0.15cm}6.13  & \textbf{2.02} &  12.40 &\hspace{0.15cm}6.51 &  \textbf{2.37}    \\
 Desk   &  11.00& \hspace{0.15cm}6.86 &   \textbf{1.25}& 11.00 & \hspace{0.15cm}7.25  & \textbf{1.70}  &  11.26  &\hspace{0.15cm}7.83 & \textbf{2.44}     \\
 Toilet  & 17.34 &13.46  & \textbf{1.81}& 17.78  & 14.55  & \textbf{2.78} &  18.25 &15.50  &\textbf{4.18}    \\
 Monitor   & 14.55 & \hspace{0.15cm}9.45 &  \textbf{1.45} & 14.72 & 10.21 & \textbf{2.05} &  14.85 &14.85 & \textbf{3.01}     \\
 Table & \hspace{0.15cm}5.95 & \hspace{0.15cm}2.63 &  \textbf{0.66} & \hspace{0.15cm}6.00 & \hspace{0.15cm}2.98 & \textbf{0.89}&  \hspace{0.15cm}6.10 & \hspace{0.15cm}3.41 &\textbf{1.21}   \\
 Night-stand &  15.01 & 12.63  & \textbf{1.76} & 16.20 & 16.26  & \textbf{3.15}& 17.74 &19.38 & \textbf{5.19}     \\
 Bathtub  & 10.18 & \hspace{0.15cm}5.39  & \textbf{1.09}&10.25   & \hspace{0.15cm}5.93   & \textbf{1.56} & 13.22 &\hspace{0.15cm}6.53&  \textbf{2.16}     \\
 Dresser&  14.65 &14.47  &  \textbf{1.41}  &  15.69 &18.33 & \textbf{2.77} & 17.57   & 21.52 &  \textbf{5.29}     \\
 \hline
 Mean Error & 11.44  & \hspace{0.15cm}8.77  &  \textbf{1.31}  & 11.94 & \hspace{0.15cm}9.37 &  \textbf{2.00} & 12.75  & 10.96 &  \textbf{3.05} \\
  \hline
\end{tabular}
\caption{\small Average error for Completion }
\end{table}
Injected slicing noise is challenging on two counts: First, our network is not trained for this noise. Secondly, injecting 30 \% of slicing noise leads to significant removal of object with large portion of object missing. hus, evaluating on this noise relates to the task of shape completion. For comparison, we again use Shapenet with the same parameters as described in the previous section.  In the Table below, 10, 20, 30 indicates the \% of slicing noise. So, 30 \% means that we randomly remove all voxels lying on 9 (30 \% ) faces of the cube. We use the same metric as described in the previous section to arrive at the following numbers in \%
\paragraph{Discussion}
Our network performance is significantly better than the CAE as well as Shapenet. This is also shown in the qualitative results shown later in Table \ref{fig:denoising}.
Our network superior performance over no noise network (CAE) justifies learning voxel occupancy from  noisy shape data. The performance on different noise also suggest that our network finds completing slicing noise (completion) more challenging than denoising random noise.  30 \% of slicing noise removes significant chunk of the object.

\section{Qualitative Comparison}

In Table \ref{fig:denoising} and \ref{fig:qualitative}, each row contains 4 images where the first one corresponds to the ground truth, second one is obtained by injecting noise (random and slicing) and acts as a input to our network. Third image is the reconstruction obtained by our network while fourth image is the outcome of shapenet.
\begin{table}[h!]
\begin{center}
\begin{tabular}{ccccc}  \\
&Reference & Noisy & Our & 3DShapeNet \\
&Model     & input & Reconstruction & Reconstruction \\\hline
&&&&\\

Chair&
\includegraphics[trim=0.2\imagewidth{} 0.1\imagewidth{} 0.2\imagewidth{} 0.05\imagewidth{}, clip, width = 0.085\imagewidth{}]{images/chair/chair_133_orig.jpg} &
\includegraphics[trim=0.2\imagewidth{} 0.1\imagewidth{} 0.2\imagewidth{} 0.05\imagewidth{}, clip, width = 0.085\imagewidth{}]{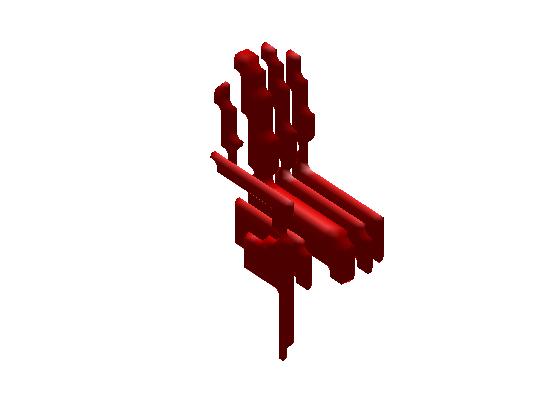} &
\includegraphics[trim=0.2\imagewidth{} 0.1\imagewidth{} 0.2\imagewidth{} 0.05\imagewidth{}, clip, width = 0.085\imagewidth{}]{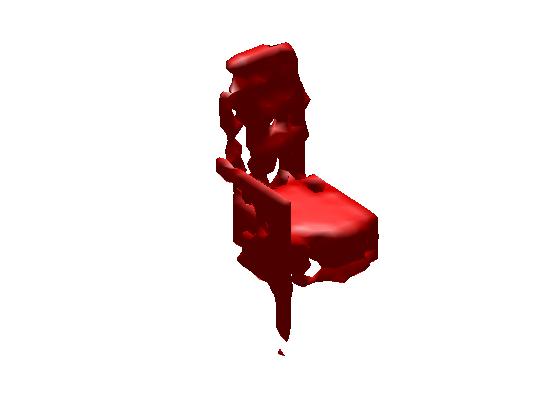} &
\includegraphics[trim=0.2\imagewidth{} 0.1\imagewidth{} 0.2\imagewidth{} 0.05\imagewidth{}, clip, width = 0.085\imagewidth{}]{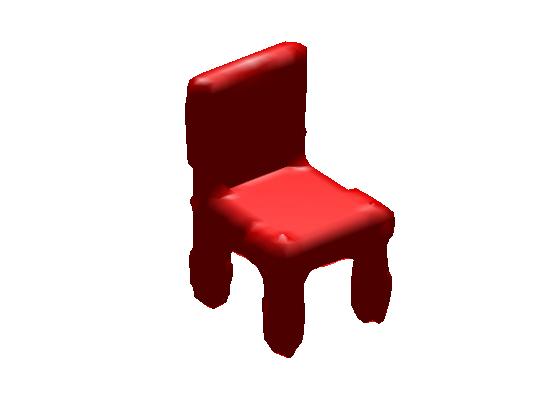} \\

Night Stand &
\includegraphics[trim=0.2\imagewidth{} 0.1\imagewidth{} 0.2\imagewidth{} 0.05\imagewidth{}, clip, width = 0.085\imagewidth{}]{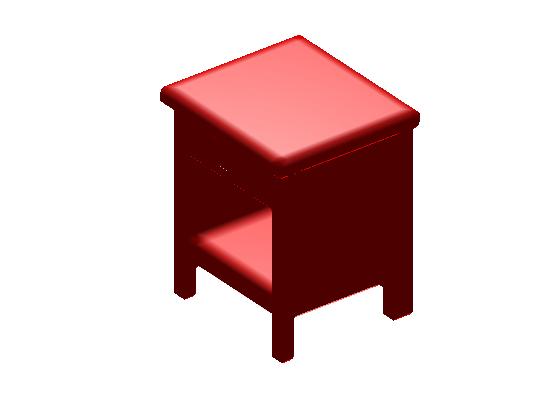} &
\includegraphics[trim=0.2\imagewidth{} 0.1\imagewidth{} 0.2\imagewidth{} 0.05\imagewidth{}, clip, width = 0.085\imagewidth{}]{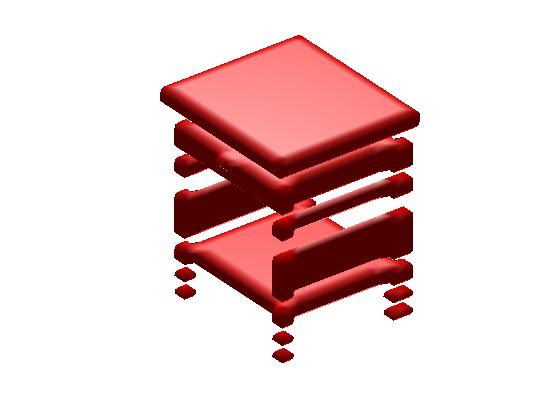} &
\includegraphics[trim=0.2\imagewidth{} 0.1\imagewidth{} 0.2\imagewidth{} 0.05\imagewidth{}, clip, width = 0.085\imagewidth{}]{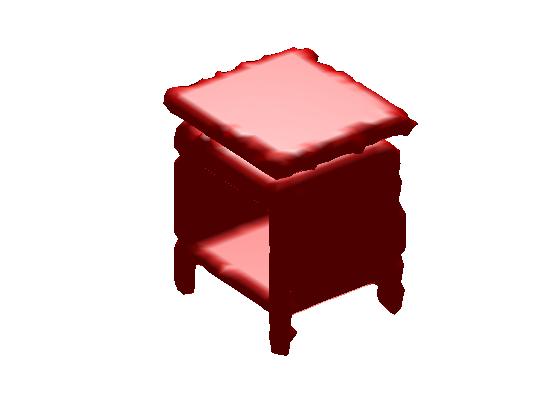} &
\includegraphics[trim=0.2\imagewidth{} 0.1\imagewidth{} 0.2\imagewidth{} 0.05\imagewidth{}, clip, width = 0.085\imagewidth{}]{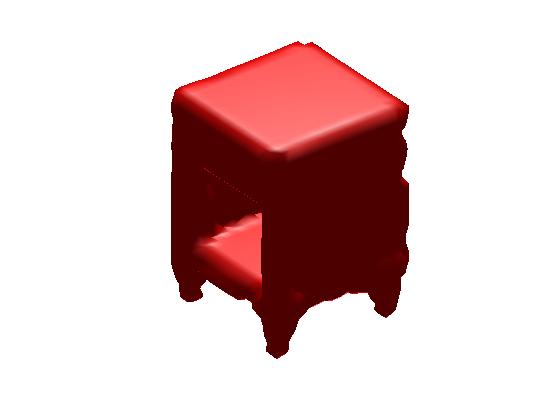} \\

Desk&
\includegraphics[trim=0.2\imagewidth{} 0.1\imagewidth{} 0.1\imagewidth{} 0.05\imagewidth{}, clip, width = 0.085\imagewidth{}]{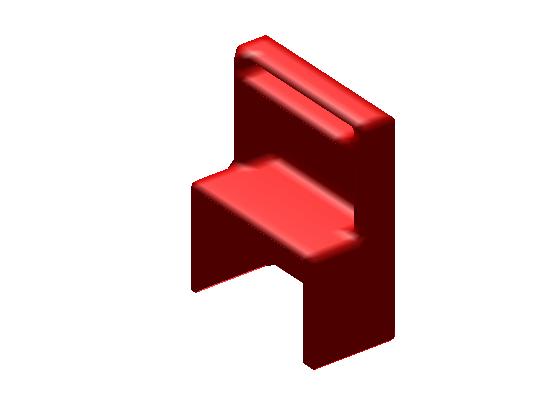} &
\includegraphics[trim=0.2\imagewidth{} 0.1\imagewidth{} 0.1\imagewidth{} 0.05\imagewidth{}, clip, width = 0.085\imagewidth{}]{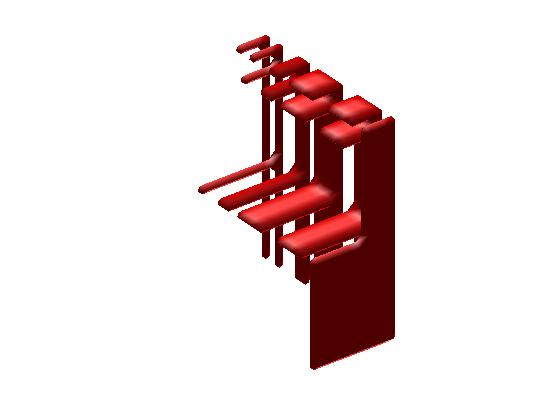} &
\includegraphics[trim=0.2\imagewidth{} 0.1\imagewidth{} 0.1\imagewidth{} 0.05\imagewidth{}, clip, width = 0.085\imagewidth{}]{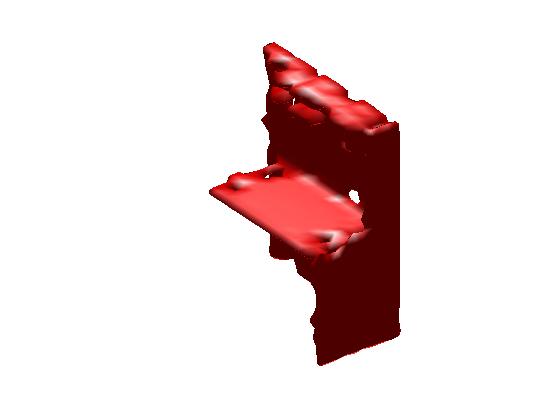} &
\includegraphics[trim=0.2\imagewidth{} 0.1\imagewidth{} 0.1\imagewidth{} 0.05\imagewidth{}, clip, width = 0.085\imagewidth{}]{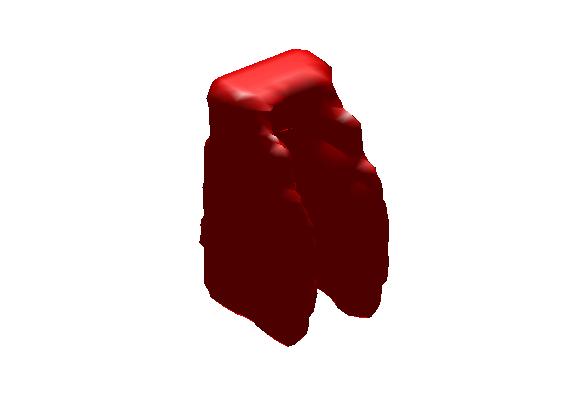} \\

Bathtub &
\includegraphics[trim=0.2\imagewidth{} 0.1\imagewidth{} 0.2\imagewidth{} 0.05\imagewidth{}, clip, width = 0.085\imagewidth{}]{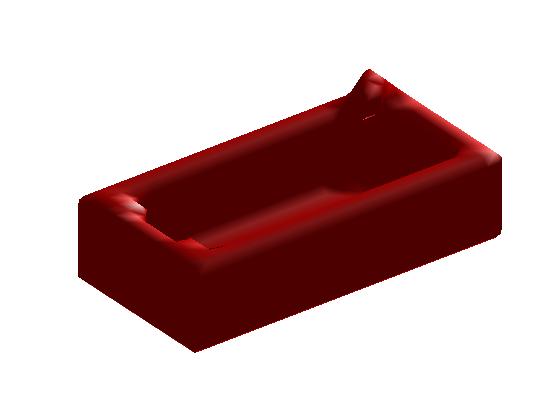} &
\includegraphics[trim=0.2\imagewidth{} 0.1\imagewidth{} 0.2\imagewidth{} 0.05\imagewidth{}, clip, width = 0.085\imagewidth{}]{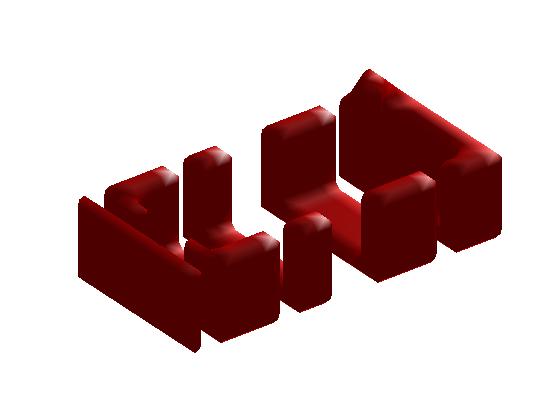} &
\includegraphics[trim=0.2\imagewidth{} 0.1\imagewidth{} 0.2\imagewidth{} 0.05\imagewidth{}, clip, width = 0.085\imagewidth{}]{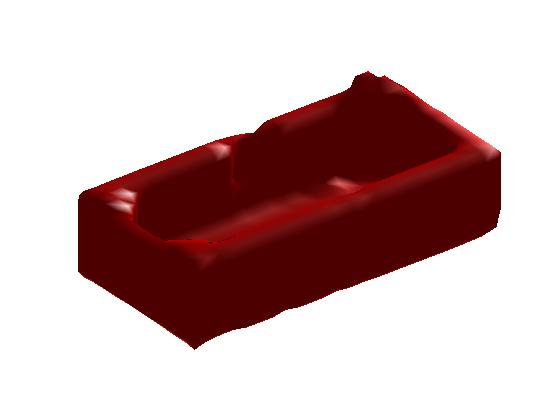} &
\includegraphics[trim=0.2\imagewidth{} 0.1\imagewidth{} 0.1\imagewidth{} 0.05\imagewidth{}, clip, width = 0.085\imagewidth{}]{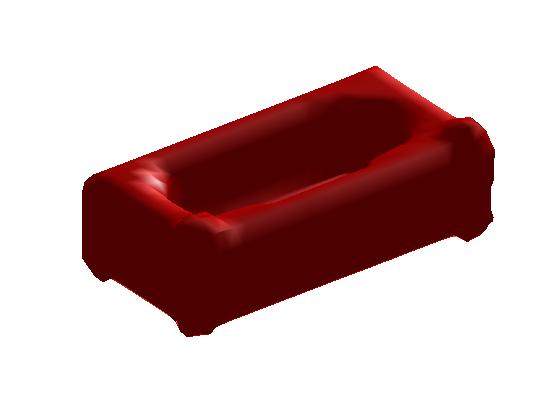} \\

Table &
\includegraphics[trim=0.2\imagewidth{} 0.1\imagewidth{} 0.2\imagewidth{} 0.03\imagewidth{}, clip, width = 0.085\imagewidth{}]{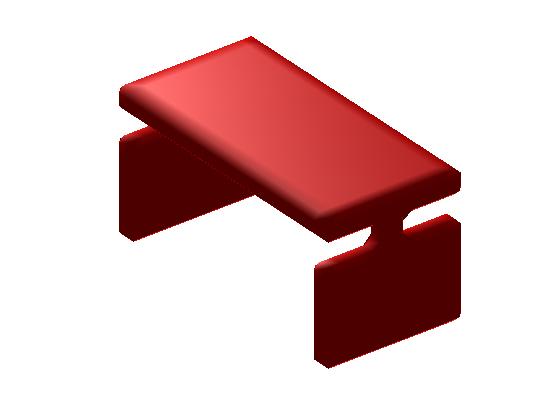} &
\includegraphics[trim=0.2\imagewidth{} 0.1\imagewidth{} 0.2\imagewidth{} 0.03\imagewidth{}, clip, width = 0.085\imagewidth{}]{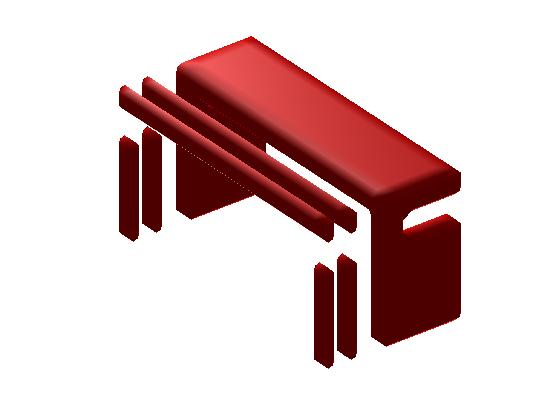} &
\includegraphics[trim=0.2\imagewidth{} 0.1\imagewidth{} 0.2\imagewidth{} 0.03\imagewidth{}, clip, width = 0.085\imagewidth{}]{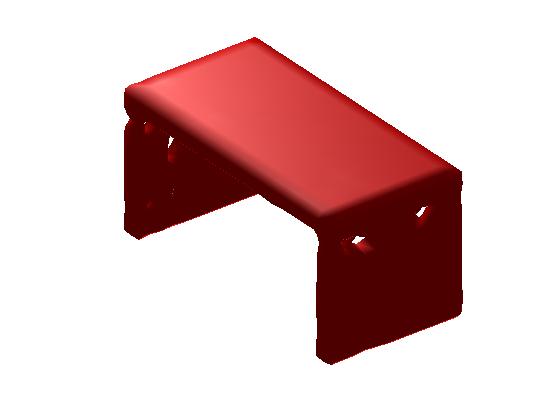} &
\includegraphics[trim=0.2\imagewidth{} 0.1\imagewidth{} 0.2\imagewidth{} 0.03\imagewidth{}, clip, width = 0.085\imagewidth{}]{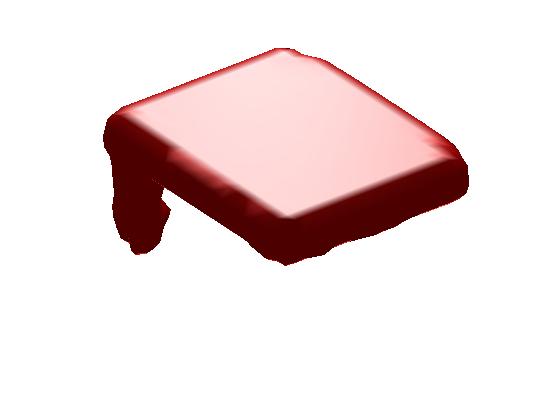}
\\
\iffalse
Sofa &
\includegraphics[trim=0.2\imagewidth{} 0.1\imagewidth{} 0.2\imagewidth{} 0.05\imagewidth{}, clip, width = 0.085\imagewidth{}]{images/sofa/sofa_43_orig.jpg} &
\includegraphics[trim=0.2\imagewidth{} 0.1\imagewidth{} 0.2\imagewidth{}
0.05\imagewidth{}, clip, width = 0.085\imagewidth{}]{images/sofa/sofa_43_dist.jpg} &
\includegraphics[trim=0.2\imagewidth{} 0.1\imagewidth{} 0.2\imagewidth{} 0.05\imagewidth{}, clip, width = 0.085\imagewidth{}]{images/sofa/sofa_43_recons.jpg} &
\includegraphics[trim=0.2\imagewidth{} 0.1\imagewidth{} 0.2\imagewidth{} 0.05\imagewidth{}, clip, width = 0.085\imagewidth{}]{images/Shapenet/sofa_43_SN.jpg} \\
\fi
Monitor &
\includegraphics[trim=0.2\imagewidth{} 0.1\imagewidth{} 0.2\imagewidth{} 0.05\imagewidth{}, clip, width = 0.085\imagewidth{}]{images/monitor/monitor_103_orig.jpg} &
\includegraphics[trim=0.2\imagewidth{} 0.1\imagewidth{} 0.2\imagewidth{} 0.05\imagewidth{}, clip, width = 0.085\imagewidth{}]{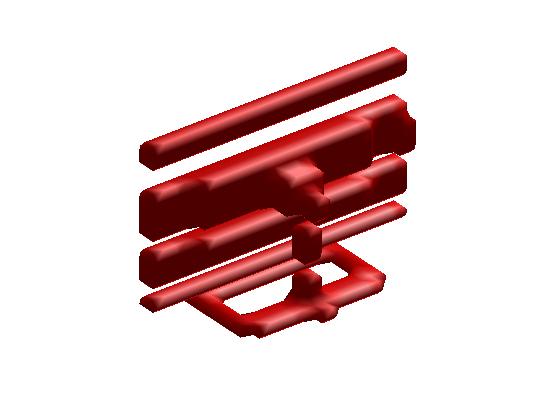} &
\includegraphics[trim=0.2\imagewidth{} 0.1\imagewidth{} 0.2\imagewidth{} 0.05\imagewidth{}, clip, width = 0.085\imagewidth{}]{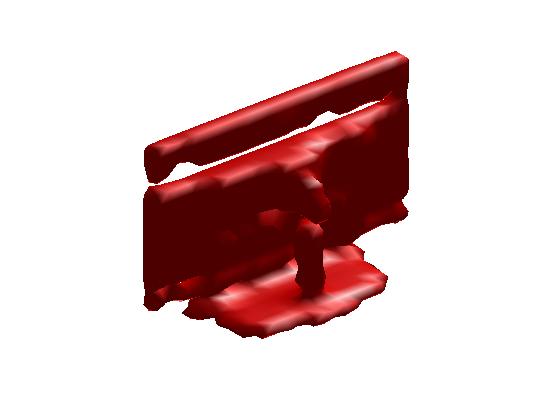} &
\includegraphics[trim=0.2\imagewidth{} 0.1\imagewidth{} 0.2\imagewidth{} 0.05\imagewidth{}, clip, width = 0.085\imagewidth{}]{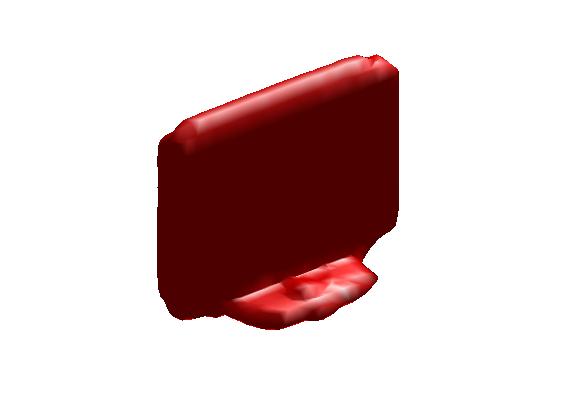} \\

Bed &
\includegraphics[trim=0.2\imagewidth{} 0.1\imagewidth{} 0.05\imagewidth{} 0.05\imagewidth{}, clip, width = 0.085\imagewidth{}]{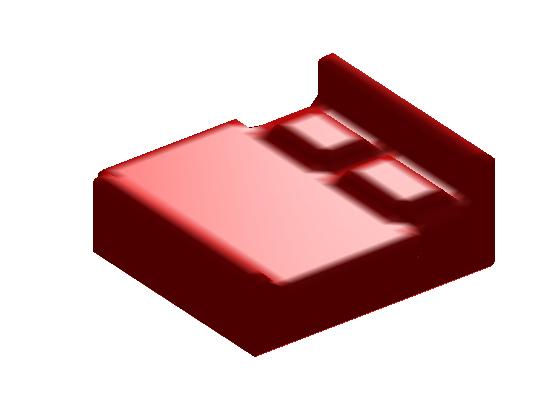} &
\includegraphics[trim=0.2\imagewidth{} 0.1\imagewidth{} 0.05\imagewidth{} 0.05\imagewidth{}, clip, width = 0.085\imagewidth{}]{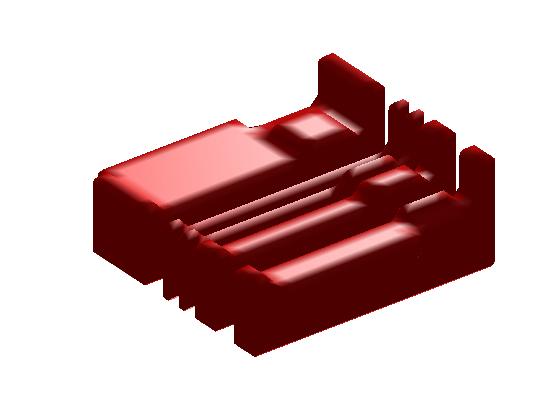} &
\includegraphics[trim=0.2\imagewidth{} 0.1\imagewidth{} 0.05\imagewidth{} 0.05\imagewidth{}, clip, width = 0.085\imagewidth{}]{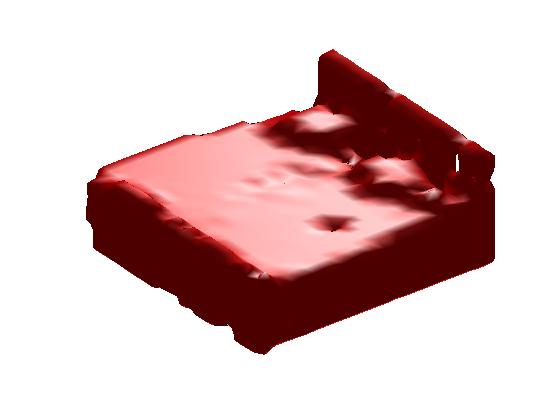} &
\includegraphics[trim=0.2\imagewidth{} 0.1\imagewidth{} 0.05\imagewidth{} 0.05\imagewidth{}, clip, width = 0.085\imagewidth{}]{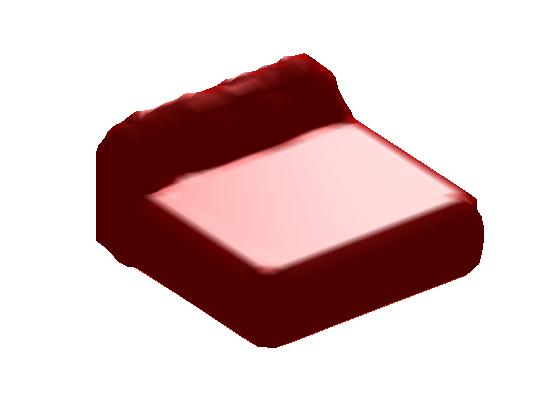} \\

\end{tabular}
\end{center}
\caption{Shape Completion results for slicing noise}\label{fig:qualitative}
\end{table}
As shown in the qualitative results, our network can fill in significant missing portion of objects when compared to shapenet. All images shown in Table \ref{fig:qualitative} are for 30 \% slicing noise scenario whereas the Table \ref{fig:denoising} corresponds to inputs with $50$\% random noise. Judging by our quantitative evaluation, our model finds slicing noise to be the  most challenging scenario. This is also evident in the qualitative results and partially explained by the fact that network is not trained for slicing noise. Edges and boundaries are smoothed out to some extent in some cases.

\paragraph{Runtime comparison with Shapenet}
We compare our runtime during train and test with Shapenet. All runtime reported here are obtained by running the code on Nvidia K40 GPU. Training time for Shapenet is quoted from their paper where it is mentioned that pre-training as well as fine tuning each takes $2$ days and test time of $600$ms is calculated by estimating the time it takes for one test completion. In contrast, our model trains in only $1$ day. We observe strongest improvements in runtime  at test time, where our model only takes $3$ms which is $200$x faster than Shapenet -- {\em an improvement of two orders of magnitude}. This is in part due to our network not requiring sampling at test time.

\section{Conclusion and Future Work}
We have presented a simple and novel unsupervised approach that learns volumetric representation by completion. The learned embedding  delivers comparable results on recognition and promising results for shape interpolation. Furthermore, we obtain  stronger results on denoising and shape completion while being trained without labels. We believe that the transition from RBM to feed forward models, first evaluation-qualitative results for shape completion, promising recognition performance and shape interpolation results will stimulate further work on deep learning for 3D geometry. In future, we plan to extend our work to deal with deformable objects and larger scenes.

\paragraph{Acknowledgement.} This work was supported by funding from the European Union's Horizon 2020 research and innovation program under the Marie\\ Sklodowska-Curie grant agreement No 642841.

\bibliographystyle{splncs}
\bibliography{kreigs}

\end{document}